\documentclass{article}

\usepackage{xcolor}
\usepackage[preprint]{corl_2026} 
\usepackage{graphicx}
\usepackage{booktabs}
\usepackage{enumitem}
\usepackage{multirow}
\usepackage{multicol}
\usepackage{cleveref}
\usepackage{caption}
\usepackage{listings}
\usepackage{tcolorbox}
\tcbuselibrary{listings}

\definecolor{codegray}{gray}{0.95}

\hypersetup{
  pdftitle={IM-ENGINE: Image Editing for Embodied Data Generation},
  pdfauthor={Yian Wang, Junyi Cao, Xiaowen Qiu, Chuang Gan},
}

\title{IM-ENGINE: Image Editing for Embodied \\ Data Generation}

\author{
  Yian Wang\\
  University of Massachusetts Amherst
  \And
  Junyi Cao\\
  University of Massachusetts Amherst
  \And
  Xiaowen Qiu\\
  Genesis AI
  \And
  Chuang Gan\\
  University of Massachusetts Amherst
}

\begin{document}
\maketitle
\vspace{-8mm}
\includegraphics[width=\linewidth]{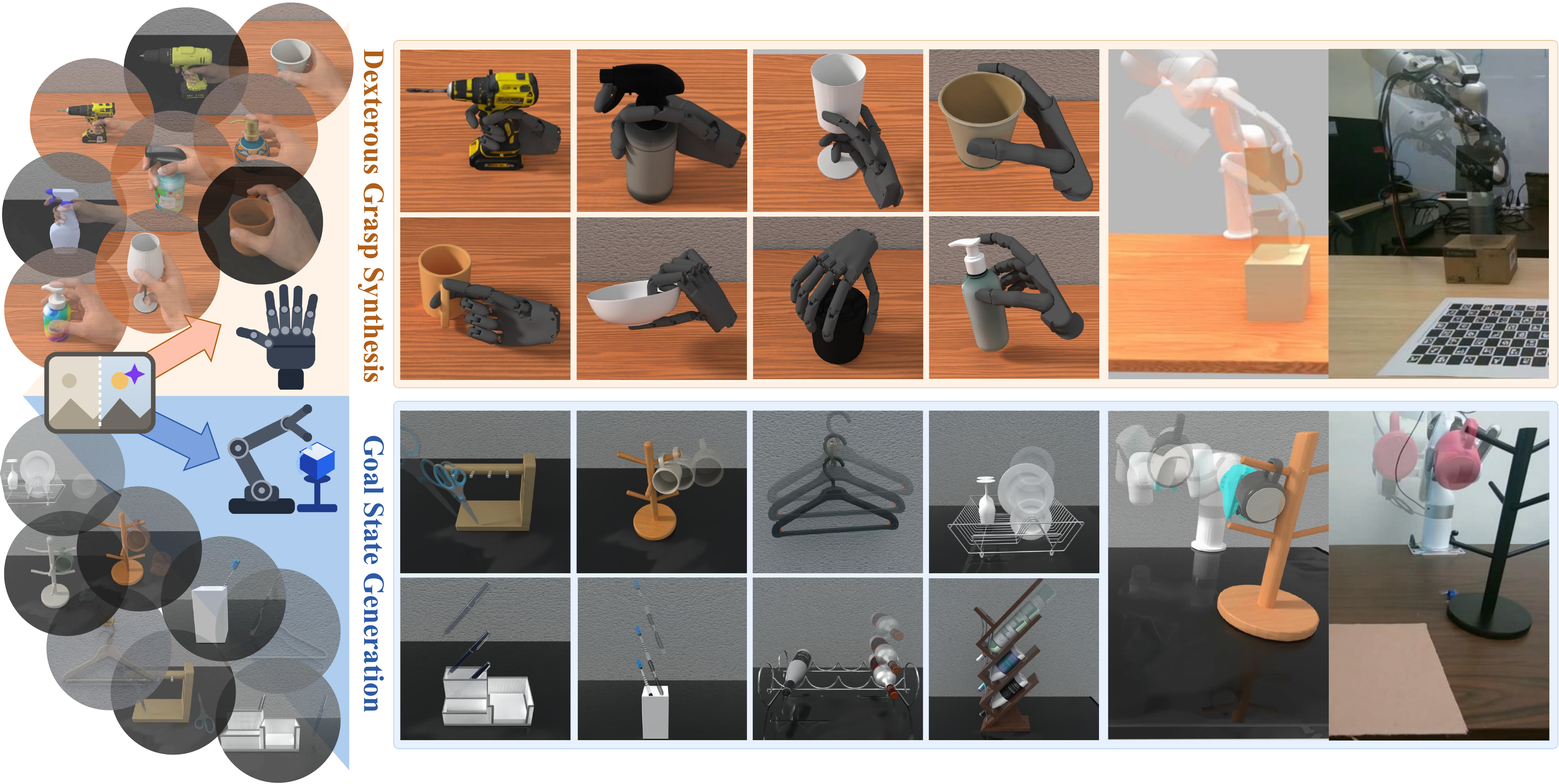}
\captionof{figure}{IM-ENGINE uses image editing as a semantic interface for simulator-grounded data generation. Edited images are grounded back into simulator state for two tasks: functional dexterous grasp synthesis and goal-state generation, yielding physically validated grasps, goals, and robot-usable trajectories.}
\label{fig:teaser}


\begin{abstract}
Learning-based manipulation requires supervision that is both semantically meaningful and physically executable, but current data pipelines often provide only one of these properties. Human demonstrations capture intent but are costly to collect and constrained by the human--robot embodiment gap, while simulation can scale data generation but often under-specifies functional behavior. We present IM-ENGINE, a simulator-grounded pipeline that uses image editing as an intermediate representation for embodied data generation. Given a rendered scene with known geometry, depth, segmentation, and camera parameters, IM-ENGINE edits the image to inject task-relevant semantics, recovers explicit 3D state using simulator priors and an unchanged anchor object, refines the state in physics, and converts it into robot-executable supervision. We instantiate the pipeline for dexterous grasp synthesis and goal-state generation. For grasping, IM-ENGINE generates a human grasp in image space, recovers the hand-object interaction, retargets it to a robot hand, and refines it into physically validated robot grasps. For goal generation, it edits a rendered scene into a desired outcome, recovers the target-object pose, and refines it into physically valid, semantically meaningful goals and trajectories. This combination of generative semantic priors and simulator grounding enables scalable task-relevant supervision for robot learning.
\end{abstract}



\section{Introduction}

Learning-based robot manipulation depends on supervision that is both semantically informative and physically executable. For many tasks, the challenge is not only to satisfy geometric constraints, but also to capture \emph{how} an object should be grasped or placed for downstream use. Human data naturally capture many of these semantic constraints, but collecting such data through teleoperation or physical demonstration remains burdensome \citep{walke2023bridgedata,khazatsky2024droid,brohan2022rt1,zitkovich2023rt2}.
Recent human-data pipelines reduce this burden through low-cost teleoperation interfaces, motion-capture systems, and wearable exoskeletons \citep{chi2024umi,wu2024gello,wang2024dexcap,shaw2025bidex,xu2025dexumi,fang2025dexop}, or leverage large-scale egocentric human videos \citep{hoque2025egodex,punamiya2025egobridge, yang2025egovla}. These approaches broaden coverage, but the supervision is still mediated by human embodiment and typically requires cross-embodiment alignment or hand-motion retargeting before it can be transferred to a target robot hand \citep{sivakumar2022robotictelekinesis,punamiya2025egobridge,li2025maniptrans,pan2025spider}.
Simulation-first pipelines can scale data generation without ongoing human involvement and expose full scene state \citep{wang2023dexgraspnet,zhong2025dexgraspanything,chen2025dexonomy,chen2024bodex,wang2023gensim,wang2023robogen,wang2024architect,sun2025threedgeneralist}, but a central challenge is to preserve task semantics rather than optimizing only for physical feasibility---for example, by placing the index finger on a spray-bottle trigger instead of simply holding the bottle from below, or by hanging the mug on a mug tree rather than merely placing it on or near a support.

This challenge is visible in grasp and goal-generation work. Large-scale synthetic dexterous grasp pipelines generate physically valid grasps at impressive scale \citep{wang2023dexgraspnet,zhong2025dexgraspanything}, but force-closure or stability objectives do not necessarily yield human-like functional grasps. Image-editing and visual-foresight methods provide semantic guidance for manipulation \citep{black2023susie,zhao2025cot,zhang2026foreact}, but usually use generated observations as policy guidance rather than explicit 3D supervision. Scene-generation work broadens object-placement semantics \citep{yang2024physcene,wang2024architect,ling2026scenethesis,wang2026physcensis,sun2025threedgeneralist,huang2025fireplace,scenesmith2026}, but often focuses on plausible layouts or simple support relations rather than intricate object-object goal states, such as slotting a plate's rim between dish-rack rails. As a result, existing pipelines still struggle to combine semantic plausibility, geometric grounding, and robot-executable structure within a single data-generation process.

To address this gap, we use image editing as an interface between semantic intent and physical grounding, as shown in Fig.~\ref{fig:teaser}. Our key observation is that editing is most useful for robotics when applied to scenes that already originate in simulation, inspired by \citet{wang2024architect}. A simulator-rendered scene provides privileged geometry, calibration, visibility, segmentation, and depth typically unavailable in unconstrained web imagery \citep{chen2025web2grasp} or directly generated images. After editing, these priors anchor reconstruction: the image supplies semantic intent, while the simulator state supports metric 3D recovery, physics verification, and conversion into supervision or trajectories.

We instantiate this idea in IM-ENGINE, a unified four-stage pipeline. Given a rendered scene of 3D assets, IM-ENGINE first edits the image to depict either a plausible human functional grasp or a desired post-manipulation goal state. It then recovers a structured state from the edit---a hand-object interaction for grasp synthesis or a 6-DoF target pose relative to a fixed anchor for goal generation. Physics validation refines this state into physically valid robot grasps or stable placements, and trajectory generation converts the validated results into robot-usable supervision, such as grasp candidates or executable placement trajectories for relation-heavy tasks like hanging a mug on a mug tree or placing a dish into a holder. Across both settings, image editing provides semantic guidance, while simulation provides geometric grounding and physical validation.

Taken together, our contributions are fourfold: (1) a simulator-grounded framework that uses image editing as an intermediate representation for embodied data generation, combining generative semantic priors with privileged simulation state; (2) perception pipelines tailored to this simulator-grounded image-editing setup, recovering explicit 3D structures---hand-object interactions for grasping and 6-DoF object poses for goal generation---rather than treating edits only as visual goals; (3) a closed loop from recovered state to robot-usable supervision through physics validation and trajectory generation; and (4) demonstrations that the same pipeline supports both functional dexterous grasp synthesis and relation-heavy goal-state generation, including hanging and fitting tasks that prior visual-goal or scene-generation pipelines do not directly address.


\section{Related Work}

\textbf{Dexterous grasp data generation.}
Simulation-first methods create large corpora of robot grasps using geometry and physics. For example, DexGraspNet~\citep{wang2023dexgraspnet} employs differentiable force-closure optimization, DexGrasp Anything~\citep{zhong2025dexgraspanything} uses a physics-aware generative model, and DRO Grasp~\citep{wei2025mathcal} utilizes a cross-embodiment distance representation. However, these approaches produce robot-centric supervision that does not capture the functional intent of grasps. Human HOI datasets~\cite{brahmbhatt2020contactpose,chao2021dexycb,taheri2020grab,yang2022oakink,jian2023affordpose} offer richer semantics but are expensive to collect and do not scale to robot-executable data. Recent methods bridge this gap: Web2Grasp~\citep{chen2025web2grasp} retargets human grasps from internet to a robot hand; language-guided grasp generation~\citep{chang2024text2grasp,cha2024text2hoi,wei2024graspasyousay,lee2026dexter} enhances functional grasping by providing detailed grasp instructions; and CorDex~\citep{he2026cordex} seeds simulation expansion from a single human demonstration. Our work pursues the same semantic goals but grounds the prior in a simulator-rendered scene with known geometry, depth, and camera parameters, so image editing injects functional intent without sacrificing scene supervision.

\textbf{Visual goal generation and spatially grounded placement.}
Prior work differs in how generated visual content enters manipulation pipelines. Visual-foresight methods use images or videos as online policy guidance rather than explicit metric goals~\citep{black2023susie, bharadhwaj2024gen2act, zhao2025cot, zhang2026foreact}: Gen2Act~\citep{bharadhwaj2024gen2act} conditions policies on generated human videos, CoT-VLA~\citep{zhao2025cot} predicts future frames before acting, and ForeAct\citep{zhang2026foreact} steers a VLA with imagined observations. Another line uses generated images or videos to recover simulation states, object motions, or trajectories \citep{qiu2025lucibot,wang2024architect,ling2026scenethesis,liang2024dreamitate,patel2025rigvid,jang2025dreamgen,dharmarajan2026dream2flow,he2026vdreamer,li2026manipdreamer3d}. LuciBot~\citep{qiu2025lucibot} generates task-completion videos and reconstructs object-pose trajectories, masks, and depth for simulation policy training. SceneThesis~\citep{ling2026scenethesis} recovers physically plausible 3D scene layouts from language and visual priors, but lacks a known simulator anchor for depth and scale alignment. In contrast, our pipeline treats an edited image as a semantic proposal, estimates a metric 6-DoF target pose in the original simulator-grounded scene, refines it in physics, and executes it with a robot.

\section{Method}
\label{sec:method}

IM-ENGINE converts image edits into robot-executable manipulation data by placing the generative model inside a simulator-grounded loop. We use world, camera, anchor-object, and target-object frames $\mathcal{W},\mathcal{C},\mathcal{A},\mathcal{O}$, and define $T_{xy}\in SE(3)$ to map points from frame $\mathcal{Y}$ to frame $\mathcal{X}$. Thus $T_{cw}$ is the world-to-camera transform, $T_{wc}=T_{cw}^{-1}$ is used for back-projection. Object poses are stored in the world frame as $T_{wa}$ for the fixed anchor and $T_{wo}$ for the target object; goal generation recovers a world-frame target pose $T_{wo}^{\star}$ whose relation to the fixed anchor is evaluated in simulation. Starting from known meshes, poses, segmentation, metric depth, camera intrinsics $K$, and pose $T_{cw}$, the simulator renders the first RGB image $s_0$ and depth map $D_0$; image editing produces $\tilde{o}$; perception recovers a structured 3D state $z$; physics validation refines it; and trajectory generation converts the result into robot supervision or executable motion.

As shown in Fig.~\ref{fig:pipeline}, we instantiate this abstraction for two manipulation problems. In \emph{Grasp Synthesis}, the anchor is the manipulated object: the object remains fixed while the edit adds a human hand performing a functional grasp. The recovered state $z$ is a robot hand grasping pose in the simulator world frame, which physics validation refines into physically valid dexterous grasps. In \emph{Goal Generation}, the anchor is a fixed container or support object: the edit moves the target object into a desired final relation, such as a mug hanging on a mug tree or a plate inserted into a rack. The recovered state $z$ is a 6-DoF target pose in the simulator world frame, evaluated by its object-object relation to the fixed anchor pose $T_{wa}$. Both settings share the same four stages: image generation, perception, physics validation, and trajectory generation.

\subsection{Image Generation}

\begin{figure}[t]
    \centering
    \includegraphics[width=1.0\textwidth]{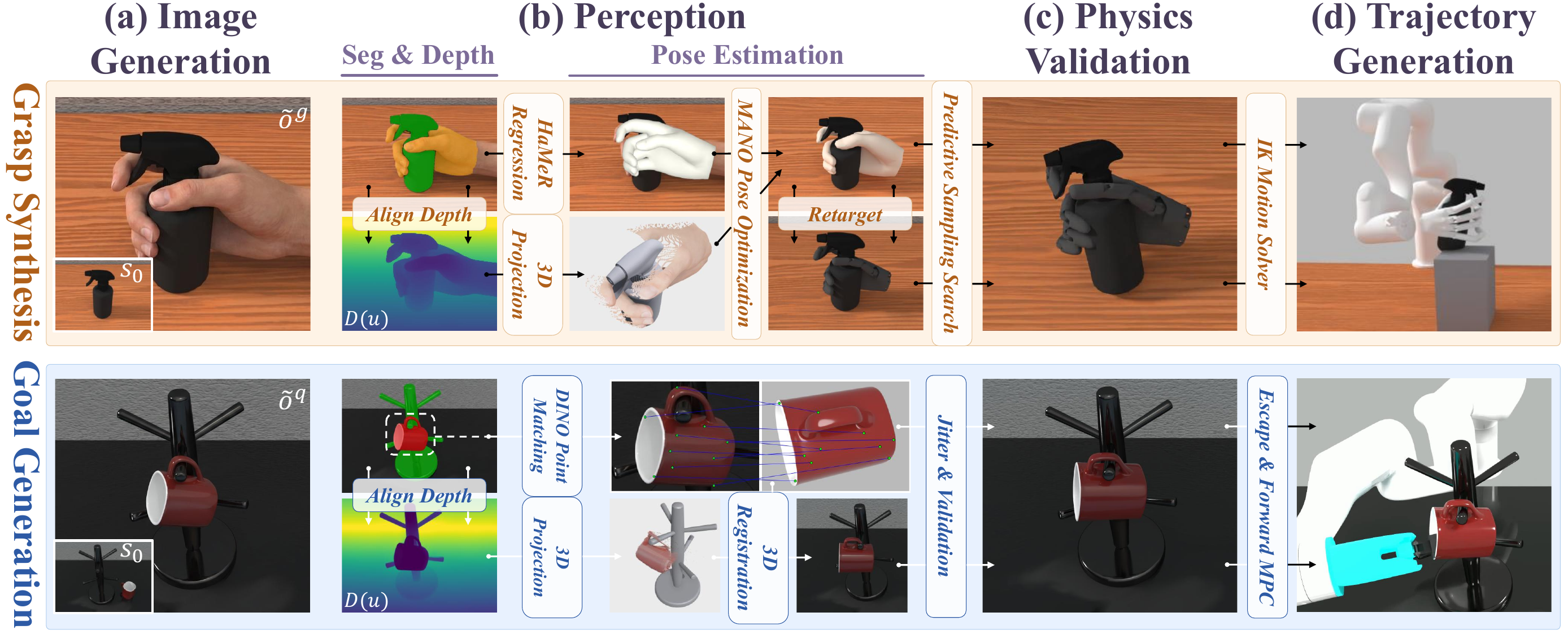}
    \caption{Our pipeline consists of four stages: (a) \emph{image generation} edits a simulator-rendered scene according to the task, producing a visual proposal for either a functional grasp or a target goal arrangement; (b) \emph{perception} lifts the edited image into a structured 3D state using simulator priors and anchor-based metric depth alignment; (c) \emph{physics validation} filters and refines the state for contact, stability, and collision feasibility; and (d) \emph{trajectory generation} converts validated states into robot manipulation data for dexterous grasp synthesis and goal-state reaching.}
    \label{fig:pipeline}
    \vspace{-6mm}
\end{figure}

The first stage creates an image-space proposal while preserving a metric reference frame (Fig.~\ref{fig:pipeline}(a)). We use \texttt{Nano Banana~2}~\citep{google2026nanobanana2} as the image-editing model. We render the initial image $s_0$ using known meshes and poses, a calibrated camera, and pixel-level segmentation. The renderer also provides a metric depth map $D_0$, so every pixel in $s_0$ can be back-projected into the simulator world frame. We keep the edited image at the same resolution as the initial render, allowing masks and depth on the unchanged anchor to be compared directly across the original and edited observations.

For \emph{Grasp Synthesis}, the scene contains a single target object on a tabletop. A vision-language model proposes functional grasp descriptions from the rendered object image, specifying contact region, grasp type, and handedness. The image editor inserts a natural human hand for each proposal while preserving the viewpoint, background, and object pose, yielding grasp-edited images, denoted generically as $\tilde{o}^{g}$, that show how the object should be grasped. For \emph{Goal Generation}, the scene contains both the target object and a fixed anchor such as a mug tree, hanger, dish rack, or holder. Rendering both objects gives the editor cues about relative scale and target shape. The editor moves the target into the requested relation while preserving the camera, object identities, and anchor pose, producing $\tilde{o}^{q}$ as a visual proposal whose fixed anchor links the edit to the simulator state.

\subsection{Perception}
\label{sec:method:perception}

Perception converts the edited observation into an explicit 3D state through segmentation, depth estimation, and pose estimation (Fig.~\ref{fig:pipeline}(b)). 

\textbf{Segmentation.} We use SAM 3~\citep{carion2025sam3} to segment the entities introduced or moved by the edit. In the grasp pipeline, we prompt for the human hand in $\tilde{o}^{g}$; in the goal pipeline, we prompt for the target object in $\tilde{o}^{q}$. The anchor mask is constructed from simulator supervision rather than relying only on recognition in the edited image: we start from the anchor mask rendered in $s_0$ and remove pixels covered by the edited foreground---the hand for grasp synthesis, or the moved target object for goal generation. The remaining pixels correspond to the visible, unchanged part of the anchor and define the alignment region. This is important because the anchor can be hard for SAM 3 to recognize when it is heavily occluded by the inserted hand or by the target object in the final goal relation.

\textbf{Depth estimation.} We estimate depth for the edited image with Depth Anything 3~\citep{lin2025depthanything3} and align the prediction to the simulator metric frame using the anchor, inspired by the depth-alignment strategy in Architect~\citep{wang2024architect}. Let $D_{\mathrm{est}}$ be the predicted depth on $\tilde{o}$, $D_0$ the renderer depth for $s_0$, and $\Omega_a$ the visible anchor region obtained from segmentation. For grasp synthesis, $a$ is the target object; for goal generation, $a$ is the fixed support or container. We solve $(\alpha, \beta) = \arg\min_{\alpha,\beta} \sum_{u \in \Omega_a}
\left( \alpha D_{\mathrm{est}}(u) + \beta - D_0(u) \right)^2$ and define the aligned depth $D(u)=\alpha D_{\mathrm{est}}(u)+\beta$. This affine alignment removes scale and offset mismatch between monocular depth and renderer depth while preserving relative geometry away from the anchor. To back-project an edited-image pixel $u$, we first compute $p_c=D(u)K^{-1}\bar{u}$ in the camera frame and then transform it to the simulator world frame as $p_w=T_{wc}p_c$, using $T_{wc}=T_{cw}^{-1}$.

\textbf{Pose estimation.} \emph{For grasp synthesis}, we recover the edited hand--object interaction by detecting the hand, estimating a MANO hand with HaMeR~\citep{pavlakos2024hamer}, refining the hand pose against the aligned hand pointcloud and known object geometry, and retargeting the refined MANO joints to the robot hand following~\cite{li2025maniptrans}. We provide the reconstruction and retargeting details in~\Cref{app:grasp_pose_estimation}.

\emph{For goal-state generation}, following \citet{ornek2024foundpose}, we recover the target object's 6-DoF pose in $\tilde{o}^{q}$ by matching DINOv3 features~\citep{simeoni2025dinov3} between rendered object templates and the edited target object, lifting depth-consistent matches into 3D, and solving a 3D registration problem to estimate the world-frame target pose $T_{wo}^{\star}$. See ~\Cref{app:goal_pose_estimation} for full template-matching and registration details.

\subsection{Physics Validation}
\label{sec:physics_validation}

The perception output is a semantic and geometric proposal, not yet an executable state. IM-ENGINE therefore uses simulation as a task-specific feasibility filter and local refinement stage before producing data (Fig.~\ref{fig:pipeline}(c)); implementation details are provided in~\Cref{app:grasp_physics_validation,app:goal_physics_validation}.

For \emph{Grasp Synthesis}, the retargeted hand initializes a parallel search over dexterous pre-grasp and post-grasp hand states. We reject candidates with initial collisions, simulate finger closing and lifting, and save grasps only if they satisfy both the lift threshold and key-fingertip contact constraints derived from the projected MANO contacts. Lifted grasps that fail the contact constraint are discarded, and the final grasp is selected by maximum total reward rather than lift height alone.

For \emph{Goal Generation}, we locally refine object poses by sampling nearby poses and simulating them with fixed anchor and environment. We reject initial collisions and accept only poses with small translation and rotation drift under gravity and contact, filtering visually plausible but physically invalid edits, \textit{e.g.}, floating placements, penetrations, and unstable hanging or insertion relations.

\subsection{Trajectory Generation}
\label{sec:traj_gen}

The final stage converts validated states into robot-executable data (Fig.~\ref{fig:pipeline}(d)). For \emph{Grasp Synthesis}, validated dexterous hand states provide grasp supervision directly, and full arm-hand rollouts are produced by planning an xArm7 approach to the selected pre-grasp, closing the dexterous hand, and lifting the object. If the default inverse-kinematics approach would disturb the object before contact, we fall back to cuRobo for a collision-aware approach. Details are in~\Cref{app:grasp_physics_validation}.

For \emph{Goal Generation}, the validated pose defines a constrained terminal relation that is difficult to reach with standard pick-and-place. We therefore compute a collision-free kinematic escape path from the goal pose using object-pose MPC, reverse it to obtain an insertion reference, and use this reference to warm-start a forward MPC search in simulation. Details are in~\Cref{app:goal_physics_validation}.

\section{Experimental Results}

\subsection{Grasp Synthesis}
\label{sec:exp_grasp_synthesis}

\paragraph{Task assets.} We consider 5 object categories in our main experiments: paper cup, power drill, spray bottle, trigger sprayer, and wine glass. For each category, we obtain 3D assets through a three-stage pipeline: text-to-image generation with \texttt{Nano Banana~2}~\citep{google2026nanobanana2}, background removal~\cite{kim2022revisiting}, and image-to-3D reconstruction~\cite{chen2025sam}. After manually filtering assets with poor geometry, we retain an average of 40 objects per category for training, and 20 more for testing. To further test robustness beyond these 5 categories, we additionally collect 50 graspable assets from BlenderKit~\citep{blenderkit}. See Fig.~\ref{fig:functional_grasping_demo} for some example results obtained by our pipeline.

\paragraph{HOI reconstruction evaluation} Before retargeting to the robot hand, we separately evaluate the recovered MANO hand--object geometry in~\Cref{app:hoi_reconstruction_eval}. The appendix compares against iHOI~\citep{ye2022whats}, EasyHOI~\citep{liu2025easyhoi}, and OakInk-Shape~\citep{yang2022oakink}, reports standard HOI reconstruction metrics, and provides qualitative analysis of how simulator grounding preserves functional contact intent.

\paragraph{Metrics.} We evaluate functional grasp generation with ShadowHand across the 5 object categories. For each grasp-generation method, we report both the lifting success rate and the functional success rate. Lifting success is evaluated in the Genesis~\cite{Genesis} physics simulator, while functional success is assessed by 5 annotators who manually inspect each generated grasp according to the criteria in~\Cref{app:appendix_func_criteria}. We report the average annotation score.

\paragraph{Baselines.} We compare against four representative methods and include ablations of DRO initialization. \textbf{DexGraspNet}~\cite{wang2023dexgraspnet} is a purely optimization-based method that requires no network inference. \textbf{DexGrasp-Anything}~\cite{zhong2025dexgraspanything} trains a diffusion model on large-scale hand poses and applies physics-guided sampling at inference to refine generated configurations. \textbf{DexGYSGrasp}~\cite{wei2024graspasyousay} is a two-stage approach that combines a language-conditioned diffusion model with a refinement stage, synthesizing task-relevant grasps. \textbf{DRO}~\cite{wei2025mathcal} predicts grasps from a configuration-invariant robot--object distance representation, enabling zero-shot transfer across hand morphologies without retraining. We evaluate two DRO variants without fine-tuning: \textbf{DRO w/o FT + r.w.}, initialized with random wrist poses, and \textbf{DRO w/o FT + g.w.}, initialized around wrist poses sampled from our pipeline. \textbf{DRO w/ FT (Ours)} denotes DRO fine-tuned on grasp data generated by IM-ENGINE.

\begin{table}
\centering
\scalebox{0.92}{
\setlength{\tabcolsep}{0.4mm}{
\begin{tabular}{l|cccccccccccc}
\toprule
\multicolumn{1}{l|}{\multirow{1}[4]{*}{Method}} & \multicolumn{2}{c}{Paper Cup} & \multicolumn{2}{c}{Power Drill} & \multicolumn{2}{c}{Spray Bottle} & \multicolumn{2}{c}{Trigger Sprayer} & \multicolumn{2}{c}{Wine Glass} & \multicolumn{2}{c}{Overall} \\
\cmidrule(lr){2-3} \cmidrule(lr){4-5} \cmidrule(lr){6-7} \cmidrule(lr){8-9} \cmidrule(lr){10-11} \cmidrule(lr){12-13}
& Lift. & Func. & Lift. & Func. & Lift. & Func. & Lift. & Func. & Lift. & Func. & Lift. & Func. \\
\midrule

DexGraspNet & 17\% &10\% & 54\% & 2\% & 71\% & 4\% & 50\% & 4\% & 36\% & 16\% & 45.6\% & 7.2\% \\
DexGrasp-Anything & \textbf{98\%} & 60\% & 98\% & 0\% & 97\% & 7\%  & 96\% & 6\% & 88\% & 14\% & 95.4\% & 17.4\% \\
DexGYSGrasp & 27\% & 21\% & \textbf{100\%} & 7\% & 84\% & 33\% & 92\% & 37\% & \textbf{100\%} & 94\% & 80.6\% & 38.4\% \\
DRO w/o FT + r.w. & 48\% & 10\% & 97\% & 2\% &  85\% & 4\% & 94\% & 6\% &90\% & 30\% & 82.8\% & 10.4\% \\
DRO w/o FT + g.w. & 73\% & 63\%  & 98\% & 11\% &  88\% & 3\% & 77\% & 6\% &90\% & 68\% & 85.2\% & 30.2\% \\
DRO w/ FT (Ours) & \textbf{98\%} & \textbf{92\%}  & \textbf{100\%} & \textbf{76\%} &  \textbf{99\%} & \textbf{69\%} & \textbf{100\%} & \textbf{81\%} &\textbf{100\%} & \textbf{98\%} & \textbf{99.4\%} & \textbf{83.2\%} \\
\bottomrule
\end{tabular}
}}
\caption{Quantitative comparison on grasp synthesis.}
\label{tab:func_grasp_exp2}
\vspace{-6mm}
\end{table}

\paragraph{Results.} \Cref{tab:func_grasp_exp2} shows that \textbf{DRO w/ FT (Ours)} achieves the best overall lifting (99.4\%) and functional success (83.2\%). The results highlight that lifting does not imply functional correctness: \textbf{DexGrasp-Anything} attains high lifting success (95.4\%) but only 17.4\% functional success, including 0\% on the power drill, because stability-driven grasps can miss task-relevant contacts. \textbf{DexGraspNet} is weakest without a learned semantic prior (45.6\% / 7.2\%), while \textbf{DexGYSGrasp} improves functional success to 38.4\% but still struggles with objects requiring precise fingertip placement, such as the power drill and spray bottle.
The DRO ablations further show the value of IM-ENGINE data. Initializing DRO with wrist poses from our pipeline (\textbf{DRO w/o FT + g.w.}) triples functional success over random initialization (30.2\% vs.\ 10.4\%), and fine-tuning on our generated data adds another 53-point gain to 83.2\%. The improvement is largest on spray bottles and trigger sprayers, where success requires placing the index finger on the trigger. We also discuss the qualitative results in~\Cref{app:grasp_visual_results}. Moreover, we report in~\Cref{app:grasp_failure_cases} that grasps generated directly by IM-ENGINE achieve approximately 86\% average functionality across the 5 categories.

\subsection{Goal Generation}
\label{sec:exp_goal_gen}

\begin{figure}[htp]
    \centering
    \includegraphics[width=\textwidth]{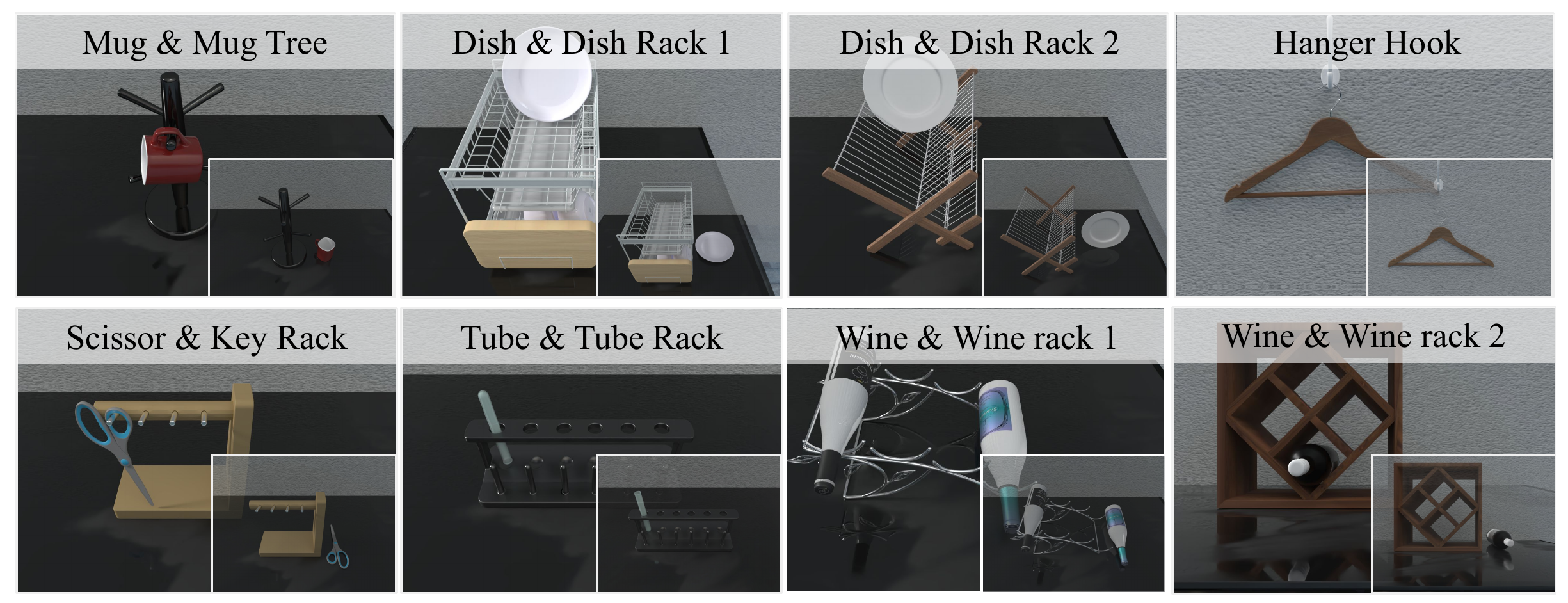}
    \caption{Qualitative results on goal-state generation. We present physically validated goal states produced by IM-ENGINE for each task, with the bottom-right images revealing the original scene.}
    \label{fig:demo_grid}
    \vspace{-6mm}
\end{figure}

\paragraph{Tasks and metrics.} We evaluate goal-state generation across eight tasks that require placing diverse objects into suitable containers, as shown in Fig.~\ref{fig:demo_grid}. A generated goal state is considered successful only if it is physically plausible in simulation---\textit{e.g.}, stable under gravity and free of penetration with the container---and the rendered final state satisfies the intended task relation.

\paragraph{Baselines.} We compare against three baselines that represent distinct strategies for goal-pose generation without image editing: 
\textbf{Molmo-2D}~\citep{deitke2025molmo} uses a VLM to click a feasible pixel in the rendered scene, lifts it to 3D with rendered depth and camera parameters, and fixes the object rotation; \textbf{VLM-6DoF} iteratively prompts a VLM with the rendered scene and current 6-DoF poses to propose an updated target pose, then validates the final pose in physics; \textbf{PhyScensis}~\citep{wang2026physcensis} applies an occupancy-grid heuristic search over candidate placements and rotations, validates them in physics, and selects the most task-consistent result with a VLM agent. 
See~\Cref{app:goal_baseline_analysis} for more details of each baseline.

\begin{table}[t]
\centering
\scalebox{0.92}{
\setlength{\tabcolsep}{1mm}{
\begin{tabular}{l|ccccccccc}
\toprule
Method & Mug & Dish 1 & Dish 2 & Hanger & Tube & Scissor & Wine 1 & Wine 2 & Overall\\
\midrule
Molmo-2D & 1/5 & 0/5 & 0/5 & 0/5 & 0/5 & 0/5 & 0/5 & 0/5 & 1/40\\
VLM-6DoF (5 iters) & 0/5 & 1/5 & 0/5 & 1/5 & 0/5 & 1/5 & 1/5 & 0/5 & 4/40\\
VLM-6DoF (10 iters) & 0/5 & 1/5 & 1/5 & 3/5 & 1/5 & 1/5  & 2/5 & 1/5 & 10/40 \\
PhyScensis & 0/5 & 3/5 & 2/5 & 0/5 & 0/5 & 0/5 & 1/5 & \bf 5/5 & 11/40 \\
Ours & \bf 5/5 & \bf 5/5 & \bf 3/5 & \bf 4/5 & \bf 5/5 & \bf 5/5 &\bf 3/5 & \bf 5/5 & \bf 35/40 \\
\bottomrule
\end{tabular}
}}
\caption{Comparison on the physical plausibility of the final goal states for goal generation.}
\vspace{-4mm}
\label{tab:goal_gen_metrics}

\end{table}

\paragraph{Results.} As shown in~\Cref{tab:goal_gen_metrics}, \textbf{Ours} achieves 35/40 successes, substantially outperforming the strongest baseline, PhyScensis (11/40). The gains are most pronounced on relation-heavy tasks such as mug hanging, tube insertion, scissors placement, and hanger placement, where the final pose must satisfy a specific object-object relation rather than merely rest stably on a support. \textbf{Molmo-2D} nearly fails entirely (1/40) because a lifted 2D click provides only a coarse translation and no reliable rotation. \textbf{VLM-6DoF} improves with more iterations, but direct metric 6-DoF reasoning remains unreliable for narrow constrained relations. \textbf{PhyScensis} succeeds on simpler placement tasks (\textit{e.g.}, wine 2 and dish 1), where occupancy-grid search can find stable poses, but fails on constrained insertion and hanging tasks (\textit{e.g.}, tube and hanger). This is because it emphasizes physical feasibility over semantic correctness, requiring unguided rotation search, roughly 40 VLM calls for filtering, and a grid discretization too coarse for contact-rich tasks such as tube insertion. In contrast, our edited-image proposal directly specifies the desired relation, allowing pose recovery and physics validation to refine a semantically meaningful goal rather than discover it from scratch.



\subsection{Real-World Experiments}

\subsubsection{Grasp Synthesis}

\begin{figure}[t]
    \centering
    \includegraphics[width=1.0\textwidth]{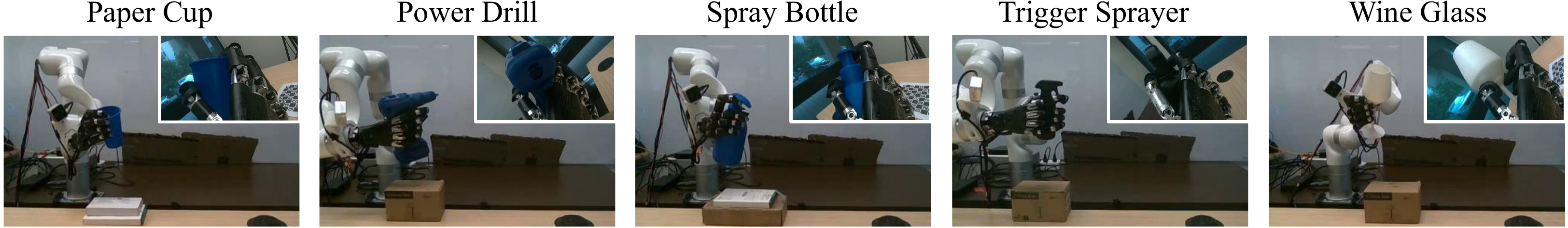}
    \caption{Real-world functional grasping demonstrations.}
    \label{fig:real-func_grasp-demo}
    \vspace{-4mm}
\end{figure}

We first execute our trajectory generation pipeline (\Cref{sec:traj_gen}) on a physical robot with generated grasps by DRO w/ FT (\Cref{sec:exp_grasp_synthesis}) across all 5 object categories. As shown in Fig.~\ref{fig:real-func_grasp-demo}, the planned arm-hand approach and grasp motions transfer to real scenes without adaptation.

\begin{figure}[t]
    \centering
    \includegraphics[width=1.0\textwidth]{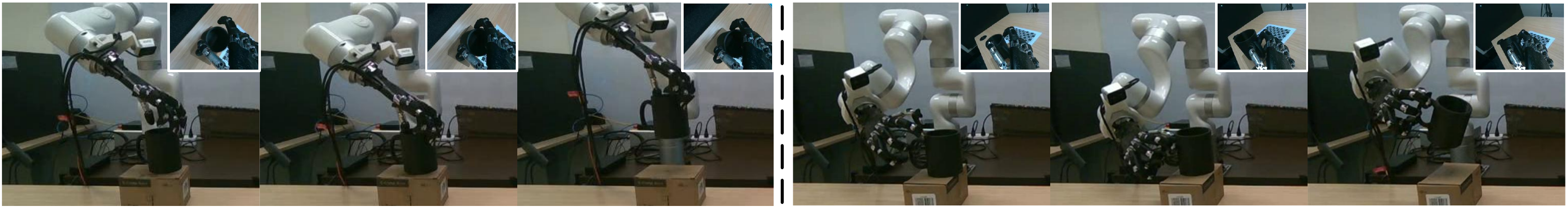}
    \caption{Real-world closed-loop deployment of two diffusion policies for mug grasping, each trained on IM-ENGINE simulation data co-trained with a small number of real demonstrations. One policy targets a top-down grasp (left) and the other targets a handle grasp (right).}
    \label{fig:real-func_grasp-policy}
    \vspace{-2mm}
\end{figure}

To test whether IM-ENGINE data supports closed-loop policy learning, we train two mug-grasp diffusion policies~\cite{chi2023diffusion}: one for top-down grasping and one for handle grasping. Each policy uses approximately 800 IM-ENGINE simulation episodes and 40 real demonstrations, with wrist-camera, front-camera, and proprioceptive observations. Concretely, we generate 5 grasp images for each grasp mode and follow~\Cref{sec:traj_gen} to obtain 160 simulation trajectories per image. We evaluate each policy over 10 trials with varied mug positions. As shown in Fig.~\ref{fig:real-func_grasp-policy}, both policies transfer to different mug instances and preserve their intended grasp modes, achieving 80\% success for top-down grasping and 60\% for handle grasping. In contrast, a diffusion policy trained with only the 40 real demonstrations achieves 0\% success. This gap indicates that IM-ENGINE-generated trajectories provide task-relevant contact structure that is difficult to learn from the limited real data alone, reducing the real-world demonstration burden for functional grasp policy learning.

\subsubsection{Goal Generation}
\label{sec:real_world_goal_gen}

\begin{figure}[htp]
    \centering
    \includegraphics[width=1.0\textwidth]{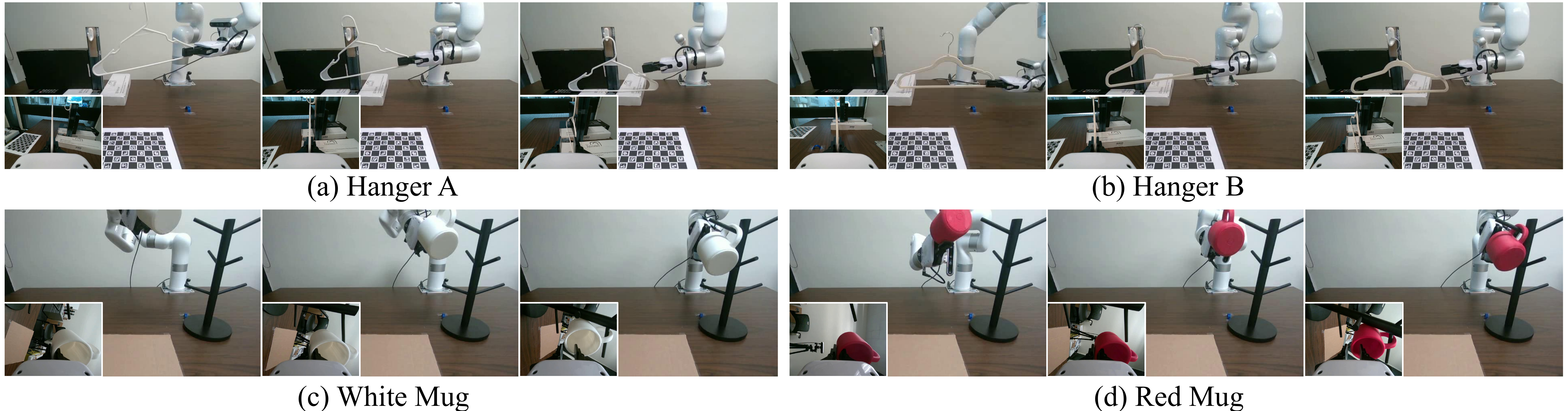}
    \caption{Real-world goal-generation results. We fine-tune SmolVLA on kinematic replays collected from simulation trajectories generated by IM-ENGINE, using both wrist camera and third-person observations as input for two separate tasks. The figure shows two successful examples for \texttt{hanger\_hook} and two successful examples for \texttt{mug\_mugtree}, each with different object instances.}
    \label{fig:real-world_goal-gen}
    \vspace{-6mm}
\end{figure}

We evaluate two representative cases, \texttt{hanger\_hook} and \texttt{mug\_mugtree}, both of which require the policy to realize a precise final object-object relationship rather than a simple placement.

To construct training data, we generate approximately 1,000 trajectories in simulation for each task by generating 10 objects from each category, using 3D object meshes generated by Meshy~\citep{Meshy}, and randomly pairing objects across the two cases. For each pair, we run our pipeline to recover a physically valid goal state. We then collect kinematic escape trajectories from the goal to diverse target poses, reverse these trajectories, define a relative grasping pose, and solve inverse kinematics to obtain robot-arm trajectory demonstrations. Finally, we replay the trajectories with randomized backgrounds and object texture, rendering both wrist-camera and third-person observations, and use the resulting data to fine-tune a SmolVLA policy~\cite{shukor2025smolvla} for each task. At deployment time, the fine-tuned policy is executed directly in the real world.

We test each policy on the corresponding task 10 times with varying object instances and initial poses; the success rates for \texttt{hanger\_hook} and \texttt{mug\_mugtree} are 70\% and 50\%, respectively. We also show two successful examples for each case in Fig.~\ref{fig:real-world_goal-gen} (see Fig.~\ref{fig:real-world_goal-gen_additional} for more quantitative results). These results indicate that simulation trajectories generated by IM-ENGINE can be converted into a policy that transfers to real scenes while preserving the desired semantic relation between the manipulated object and the anchor structure.


\section{Conclusion and Limitations}

We presented IM-ENGINE, a simulator-grounded data-generation framework that uses image editing as an interface between semantic intent and robot-executable structure. Rather than treating generated images as final visual goals, IM-ENGINE edits simulator-rendered scenes and grounds the edits back into explicit 3D state using known geometry, segmentation, depth, and physics. This lets generative models specify functional intent while simulation provides metric reconstruction, physical validation, and trajectory supervision. We instantiated IM-ENGINE for functional dexterous grasp synthesis, where edited human grasps are recovered and retargeted into robot grasp data, and goal-state generation, where edited scene outcomes become stable 6-DoF placements and executable trajectories. Experiments show improved functional grasp success with physical feasibility, strong performance over direct visual or geometric goal-generation baselines on relation-heavy tasks, and simulation trajectories that transfer to real-world policy execution.

\paragraph{Limitations.}
IM-ENGINE inherits limitations from both image editing and simulation. Edited images can specify infeasible interactions, alter anchor objects, or miss functional contacts, while collision handling and contact-rich simulation remain challenging for thin structures and constrained insertion or hanging tasks. The pipeline also assumes suitable paired 3D assets for goal generation, and grasp validation can occasionally favor stable but semantically incorrect contacts. We provide more detailed analyses in \Cref{app:grasp_failure_cases,app:goal_failure_cases}.


\clearpage


\bibliography{example}  


\clearpage
\appendix

\section{Grasp Synthesis}
\label{app:functional_grasping}

\subsection{HOI Reconstruction Evaluation}
\label{app:hoi_reconstruction_eval}
\label{app:easyhoi}

Before retargeting the edited human grasp to a robot hand, we evaluate the recovered MANO hand-object geometry. This experiment isolates the perception component of IM-ENGINE by comparing our simulator-grounded reconstruction against HOI methods that do not use simulator priors such as object pose, metric depth, or object masks. We compare against the following baselines and reference data:
\begin{itemize}[leftmargin=*]
    \item \textbf{iHOI}~\citep{ye2022whats}: a single-image HOI method that predicts a hand pose and hand-centric implicit object shape. Following Web2Grasp~\citep{chen2025web2grasp}, we align the ground-truth object mesh to this shape with ICP before computing metrics, using checkpoints fine-tuned on MOW and HO3D.
    \item \textbf{EasyHOI}~\citep{liu2025easyhoi}: a single-image reconstruction pipeline that combines foundation models for hand pose, object shape, segmentation, and inpainting, followed by prior-guided optimization. We describe our baseline adaptation below.
    \item \textbf{OakInk-Shape}~\citep{yang2022oakink}: a dataset reference rather than a reconstruction baseline. Its interactions come from real demonstrations with object poses tracked by synchronized MoCap and MANO hand poses optimized from multi-view 2D keypoints, then transferred to related object shapes; we use it as an approximate ground-truth scale for these metrics.
\end{itemize}

\begin{table}[htp]
\centering
\small
\setlength{\tabcolsep}{4pt}
\begin{tabular}{lcccc}
\toprule
Method & PDM$_{\max}$ [cm] $\downarrow$ & PV [cm$^3$] $\downarrow$ & DD [cm] $\downarrow$ & CR [\%] $\uparrow$\\
\midrule
OakInk-Shape & 0.010 & 0.050 & 2.820 & 6.9\\
\midrule
iHOI (MOW FT) & 0.588 & 23.248 & 2.457 & \textbf{23.9}\\
iHOI (HO3D FT) & 0.474 & 11.616 & 2.979 & 17.3\\
EasyHOI & 0.476 & 6.566 & \textbf{2.420} & 11.9\\
Ours (KIT) & \textbf{0.393} & \textbf{5.225} & 3.400 & 10.8\\
\bottomrule
\end{tabular}
\caption{HOI reconstruction quality comparison on our generated cases. We report maximum penetration depth (PDM$_{\max}$), penetration volume (PV), disjoint distance (DD), and contact region ratio (CR). Lower is better except for CR. MOW FT and HO3D FT denote the datasets used to fine-tune the iHOI baseline checkpoints.}
\label{tab:hoi_metrics}
\end{table}

\paragraph{EasyHOI baseline adaptation.}
We follow EasyHOI's hand-object pose-optimization pipeline~\citep{liu2025easyhoi} as a baseline, with three changes:
\begin{itemize}[leftmargin=*]
    \item \textbf{Skip mesh generation.} EasyHOI normally reconstructs the object mesh from the image; we replace this stage with the ground-truth GLB from the kit, so the optimization aligns the hand against a known object.
    \item \textbf{Use HaMeR initialization.} Instead of HandOccNet, we run HaMeR~\citep{pavlakos2024hamer} on the input image to obtain the initial MANO pose, shape, and camera, then place the hand in the kit camera frame by back-projecting the ground-truth hand-mask centroid at a depth calibrated so the rendered silhouette area matches the ground-truth hand mask.
    \item \textbf{Use ground-truth masks.} EasyHOI relies on predicted hand/object masks; we substitute the kit's \texttt{mask\_hand\_grasp}, \texttt{mask\_object\_grasp}, and \texttt{mask\_object\_render}, making the silhouette IoU loss and mask-derived object-contact points for stage-2 ICP noise-free.
\end{itemize}

\begin{figure}[t]
    \centering
    \includegraphics[width=\textwidth]{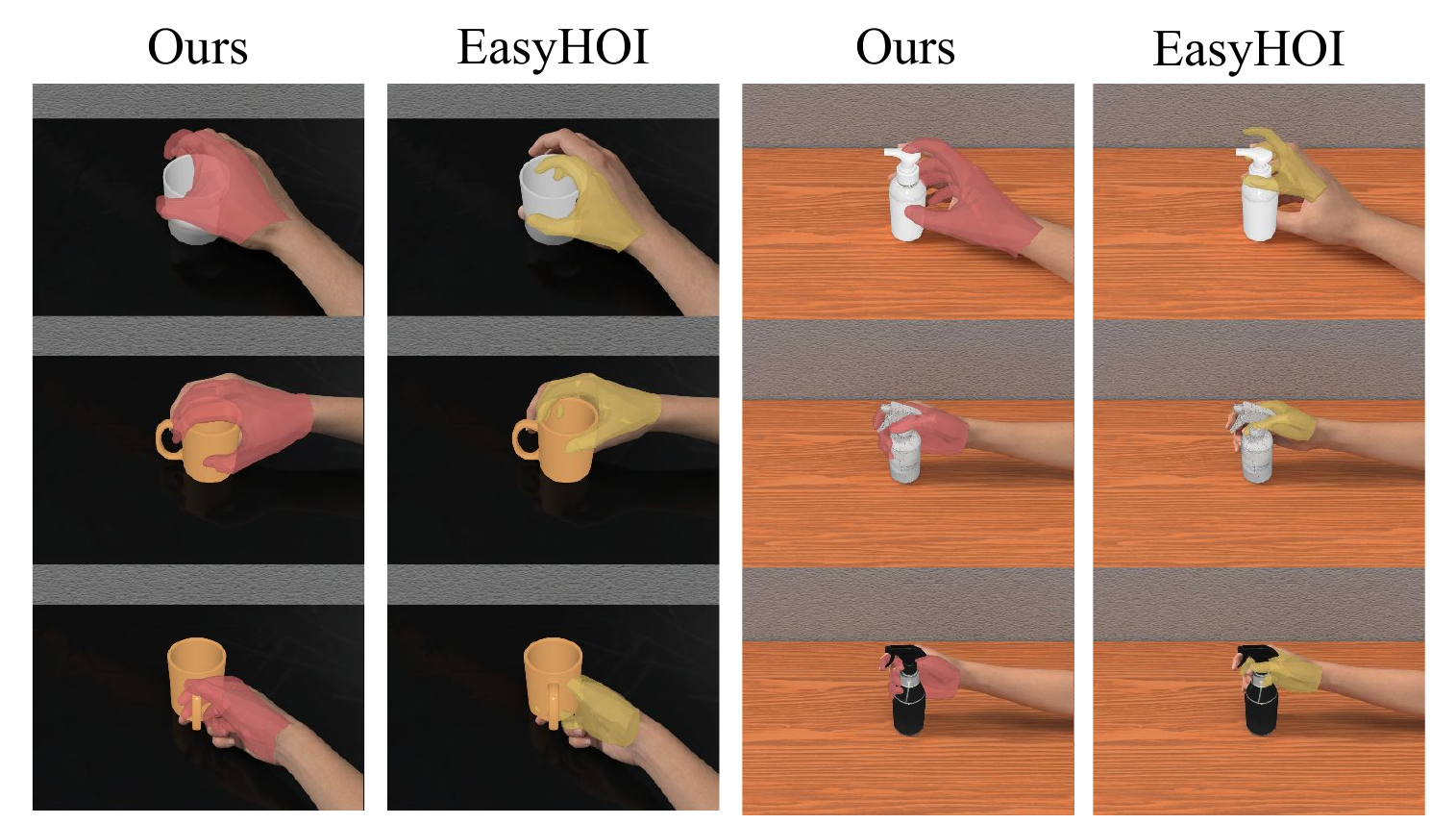}
    \caption{Qualitative HOI reconstruction comparison between our method and the adapted EasyHOI baseline.}
    \label{fig:compare_vs_hoi}
\end{figure}

\paragraph{Analysis.}
We report four standard HOI metrics: maximum penetration depth (PDM$_{\max}$), penetration volume (PV), disjoint distance (DD), and contact region ratio (CR). Lower is better except for CR. Our reconstruction substantially reduces penetration compared with the iHOI checkpoints and the adapted EasyHOI baseline. Although the EasyHOI baseline obtains similar aggregate scores to ours, it does not always preserve the functional intention of the edited grasp. While iHOI and EasyHOI can obtain lower DD or higher CR, these scores can be inflated by hand--object interpenetration rather than physically plausible contact. In contrast, our CR is closer to the OakInk-Shape reference while maintaining much lower penetration, indicating a more reasonable contact level.

As shown in Fig.~\ref{fig:functional_grasping_demo}, although the recovered hand--object geometry can occasionally contain small penetrations, it largely preserves the intended functional grasp mode. These residual penetrations can be mitigated during the post-retargeting physics-validation step, which refines the robot hand--object configuration while preserving task-relevant contacts and still produces reasonable grasps, as in the teacup case. Fig.~\ref{fig:compare_vs_hoi} further shows that the adapted EasyHOI baseline can deviate from the functional intention implied by the edited images. For example, the EasyHOI reconstruction sometimes places the index finger far from the trigger sprayer's trigger or fails to grip the cup handle precisely. In contrast, our reconstruction more closely resembles the edited images' intended grasp mode, preserving task-relevant contacts before retargeting.

To further evaluate the pipeline before physics validation, we directly test whether the retargeted robot pose already forms a firm grasp. We initialize the robot hand in a fully open pose, command it to the retargeted configuration, and then evaluate whether the resulting grasp can lift the object. Even without the physics-validation stage, this direct-retargeting test succeeds at 57\%, 98\%, 94\%, 98\%, and 81\% for paper cup, power drill, spray bottle, trigger sprayer, and wine glass, respectively.

\subsection{Method Details}
\label{app:grasp_physics_validation}

\paragraph{Functional grasp image-editing prompts.}
For functional grasp image editing, we use the following prompt template:
\begin{lstlisting}[basicstyle=\ttfamily\small,breaklines=true]
Please edit this image to show {grasp description}. 
- Keep the same camera viewpoint, lighting, and background. 
- Make the hand natural and physically plausible. 
- Show proper finger placement. 
- Keep the object in the same position. 
- Do not add objects or change the scene.
\end{lstlisting}
The category-specific grasp descriptions are:
\begin{lstlisting}[basicstyle=\ttfamily\small,breaklines=true]
paper cup: a right hand grasping the cup from the side, all four fingers wrapping around the cup body, and the thumb on the opposite side
wine glass: a right hand grasping the stem just below the bowl, with the thumb on one side, the index and middle fingers around the opposite side, the ring finger lightly supporting the lower stem, the wrist vertical, and the glass upright
power drill: a right hand holding the handle in a pistol grip, with the index finger on the trigger and the other fingers around the grip
spray bottle: a right hand grasping the bottle with the index finger pressing the pump button
trigger sprayer: a right hand grasping the sprayer with the index finger pulling the trigger
\end{lstlisting}

\paragraph{View selection.}
When rendering the first image, the renderer uses a fixed camera direction for every object, with yaw $180^\circ$ and pitch $45^\circ$, corresponding to a 3/4 elevated front view looking along the negative $x$ direction. The camera looks at the center of the object AABB, and automatically adjusts its distance so the projected scene occupies a consistent fraction of the image. Specifically, the renderer computes horizontal and tilt-aware vertical scene extents, solves the distances needed to fit each extent under a fill ratio of 0.425 of the whole image, and uses the larger distance to avoid cropping. We render with a $50^\circ$ field of view at $1200\times896$ resolution, matching the output resolution of \texttt{Nano Banana 2}, and save the resulting camera intrinsics and poses for downstream reconstruction. This view selection process work the same in Goal Generation.

\paragraph{Grasp-pose estimation.}
\label{app:grasp_pose_estimation}
Given an edited grasp image, we first localize the hand crop and run HaMeR~\citep{pavlakos2024hamer} to obtain an initial MANO mesh and 21 joints. The HaMeR prediction is then grounded in the simulator frame by optimizing a rigid transform $T_{\Delta}\in SE(3)$ from the initial HaMeR frame to the world frame. Let $\mathcal{V}_{h}^{\mathrm{vis}}$ be the visible MANO vertices, $\mathcal{V}_{h}^{\mathrm{all}}$ all MANO vertices, $\mathcal{P}_{h}$ the aligned world-frame hand point cloud from the edited image, $\mathcal{S}_{o}$ the known object surface in the world frame, and $\phi_o$ the corresponding world-frame signed-distance field. We minimize
\begin{equation}
\mathcal{L}_{\mathrm{hand}}(T_{\Delta}) =
\lambda_{\mathrm{ch}}\mathcal{L}_{\mathrm{ch}} +
\lambda_{\mathrm{sdf}}\mathcal{L}_{\mathrm{sdf}} +
\lambda_{\mathrm{con}}\mathcal{L}_{\mathrm{con}},
\end{equation}
where the visible-surface term aligns the transformed hand mesh to the observed hand points,
\begin{equation}
\mathcal{L}_{\mathrm{ch}} =
\frac{1}{|\mathcal{V}_{h}^{\mathrm{vis}}|}\sum_{v\in\mathcal{V}_{h}^{\mathrm{vis}}}\min_{p\in\mathcal{P}_{h}}\left\|T_{\Delta}v-p\right\|_2^2+
\frac{1}{|\mathcal{P}_{h}|}\sum_{p\in\mathcal{P}_{h}}\min_{v\in\mathcal{V}_{h}^{\mathrm{vis}}}\left\|p-T_{\Delta}v\right\|_2^2,
\end{equation}
the penetration term discourages hand vertices from entering the object,
\begin{equation}
\mathcal{L}_{\mathrm{sdf}}=\frac{1}{|\mathcal{V}_{h}^{\mathrm{all}}|}\sum_{v\in\mathcal{V}_{h}^{\mathrm{all}}}\max\left(0,-\phi_o(T_{\Delta}v)\right)^2,
\end{equation}
and the contact term attracts near-contact hand vertices to the object surface,
\begin{equation}
\mathcal{L}_{\mathrm{con}}=\frac{1}{|\mathcal{C}_{h}|}\sum_{v\in\mathcal{C}_{h}}\min_{s\in\mathcal{S}_{o}}\left\|T_{\Delta}v-s\right\|_2^2,
\end{equation}
where $\mathcal{C}_{h}$ denotes the predicted or selected near-contact hand vertices. After refinement, the transformed MANO joints $T_{\Delta}J_h$ are retargeted to the robot hand by solving for robot wrist pose and joint configuration $q$ that match corresponding MANO joints and fingertips under robot forward kinematics,
\begin{equation}
(T_{wr}^{\star},q^{\star})=\arg\min_{T_{wr},q}\sum_k \rho_k\left\|T_{wr}\,\mathrm{FK}_k(q)-T_{\Delta}J_{h,k}\right\|_2^2,
\end{equation}
where $T_{wr}$ maps the robot wrist frame to the world frame and $\mathrm{FK}_k(q)$ returns the $k$th robot keypoint in the wrist frame. Joint limits are enforced by the robot model. The resulting wrist pose and hand configuration initialize the physics-validation search below.

\paragraph{Physics validation.}
For grasp synthesis, physics validation uses the Genesis simulator~\citep{Genesis} to perform a parallel stochastic search over a two-frame dexterous action initialized from the retargeted hand. Because Genesis can step many environments in parallel on the GPU, this validation remains fast even though each iteration evaluates physics rollouts for many grasp candidates. Each candidate is a 50-dimensional vector
\[
\mathbf{x}=\left[\mathbf{p}_{w},\theta_{w},q_{\mathrm{pre}},\Delta q\right],
\]
where $\mathbf{p}_{w}\in\mathbf{R}^{3}$ and $\theta_{w}\in\mathbf{R}^{3}$ are the wrist position and Euler angles, $q_{\mathrm{pre}}\in\mathbf{R}^{22}$ is a collision-free pre-grasp hand configuration clipped to joint limits, and $\Delta q\in\mathbf{R}^{22}$ is a per-joint closing delta clipped to the ShadowHand closing direction, with fixed joints forced to zero. At each iteration, environment 0 evaluates the current mean exactly and the remaining environments sample Gaussian perturbations around it. The post-grasp command is $q_{\mathrm{post}}=q_{\mathrm{pre}}+\Delta q$.

The reward combines physics success, perception fidelity, and contact guidance:
\[
R(\mathbf{x})=0.30 C+0.20 L+\lambda_{\mathrm{per}}\left(0.10 r_{2\mathrm{D}}+0.02 r_{\mathrm{ori}}+0.02 r_{\mathrm{pos}}+0.10 r_{\mathrm{dof}}+0.10 r_{3\mathrm{D}}\right)+w_{c}r_{c}.
\]
Here $C$ is a binary reward for a collision-free pre-grasp pose and $L$ is a binary reward for a successful lift, defined by $C=1$ and $z_{\mathrm{lift}}\geq0.15\,\mathrm{m}$ after a 10-waypoint, 0.2 m lifting motion. The perception scale is $\lambda_{\mathrm{per}}=0.5$ by default, so collision and lift terms dominate the search. The perception terms encourage agreement with the recovered interaction: $r_{2\mathrm{D}}$ matches fingertip reprojections to HaMeR keypoints with a linear clamp from 30 px to 200 px followed by a square root, $r_{\mathrm{ori}}=\exp(-2\alpha)$ matches wrist orientation, $r_{\mathrm{pos}}=\exp(-20e_{p})$ matches wrist position, $r_{\mathrm{dof}}=\exp(-6e_{q})$ keeps the hand close to the retargeted pose, and $r_{3\mathrm{D}}$ matches 3D fingertip positions with a 1--4 cm clamped score. Thumb and index terms are weighted more strongly in the 2D and contact objectives.

Contact guidance scores whether each fingertip touches the object near its MANO-predicted contact point. For fingertip $f$ with contact distance $d_f$ and weight $a_f$, we use
\[
r_{c}=\frac{\sum_f a_f r_f}{\sum_f a_f},\qquad
r_f=\left\{\begin{array}{ll}
\max(0,1-d_f/0.03), & \mathrm{if\ fingertip\ } f \mathrm{\ contacts},\\
-0.1, & \mathrm{otherwise}.
\end{array}\right.
\]
The default contact weight is $w_c=0.20$ and the default fingertip weights are $[3,3,1,1,1]$ for thumb, index, middle, ring, and little fingers; for spray-bottle-like tasks we further emphasize the index finger.

Filtering is enforced as hard constraints rather than only through reward. First, candidates colliding at the pre-grasp pose are teleported away during rollout, preventing false lift successes. Second, a lift mask requires the collision-free hand to raise the object by at least 0.15 m. Third, the contact constraint requires all three key fingertips---thumb, index, and middle---to touch within a maximum distance. A candidate is saved only if
\[
S(\mathbf{x})=C\;\wedge\;(z_{\mathrm{lift}}\geq0.15\,\mathrm{m})\;\wedge\;K,
\]
where $K$ denotes the contact constraint. Grasps that lift but fail $K$ are counted and discarded. The next mean is the reward-weighted average of the top elites,
\[
\mu_{t+1}=\frac{\sum_{i\in E} R_i\mathbf{x}_i}{\sum_{i\in E} R_i},
\]
with annealed sampling noise. Across all banked successes, the final grasp is selected by maximum total reward rather than by lift height alone.

\paragraph{Trajectory generation.}
Validated dexterous hand states can be used directly as supervision for grasp learning. To produce full arm-hand rollouts, we mount the dexterous hand on an xArm7 and plan a Cartesian approach to the selected pre-grasp. A chain-seeded inverse-kinematics solver tracks a piecewise-linear end-effector path with orientation interpolation from a home pose to a stand-off pose and then to the pre-grasp. During execution, the arm tracks the planned path, the fingers interpolate to the pre-grasp configuration, close to the validated grasp target, hold to settle contacts, and lift while maintaining the grasp. If the inverse-kinematics tracker produces an approach that contacts or knocks the object before the grasp, we fall back to a cuRobo motion planner to generate a collision-aware approach to the same pre-grasp.

\subsection{Dual-Hand Grasping}
\label{app:dual_hand_grasping}

Although our main experiments primarily use right-hand grasps, the pipeline does not impose a right-hand-only or single-hand restriction. The image-editing prompt can specify a left hand or two hands, and the recovered interaction can be grounded in the same simulator frame before physics validation. For dual-hand settings, the validation objective can be applied jointly to two wrist poses and two hand configurations while checking contacts and lift stability on the shared object.

\begin{figure}[htp]
    \centering
    \includegraphics[width=\textwidth]{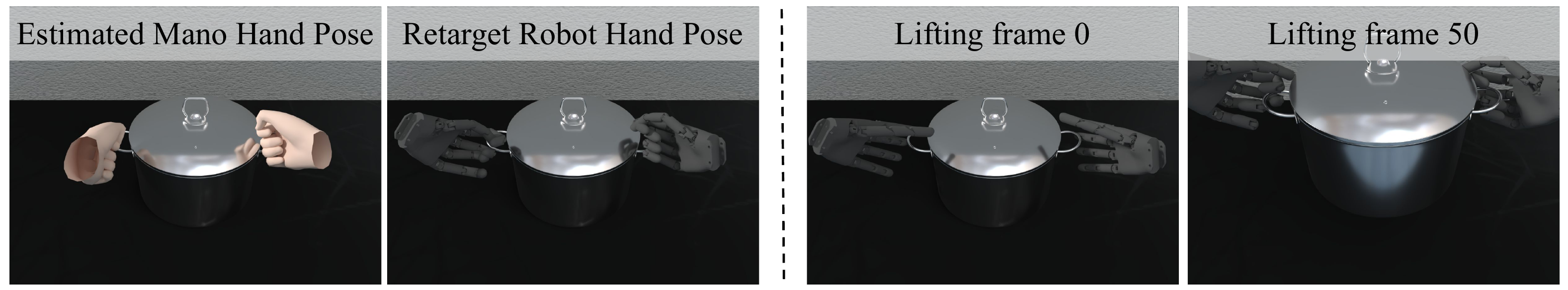}
    \caption{Example dual-hand grasp generated by IM-ENGINE for lifting a pot by grasping its two handles.}
    \label{fig:dual_hand_grasping}
\end{figure}

Fig.~\ref{fig:dual_hand_grasping} illustrates this extension on a pot-lifting task. We edited the image to specify two hands grasping the two handles, giving the downstream reconstruction and validation stages explicit semantic targets for a coordinated lift. This example highlights that our pipeline can generate functional bimanual intent beyond the single-hand grasps emphasized in our quantitative experiments.

\begin{figure}[t]
    \centering
    \includegraphics[width=\textwidth]{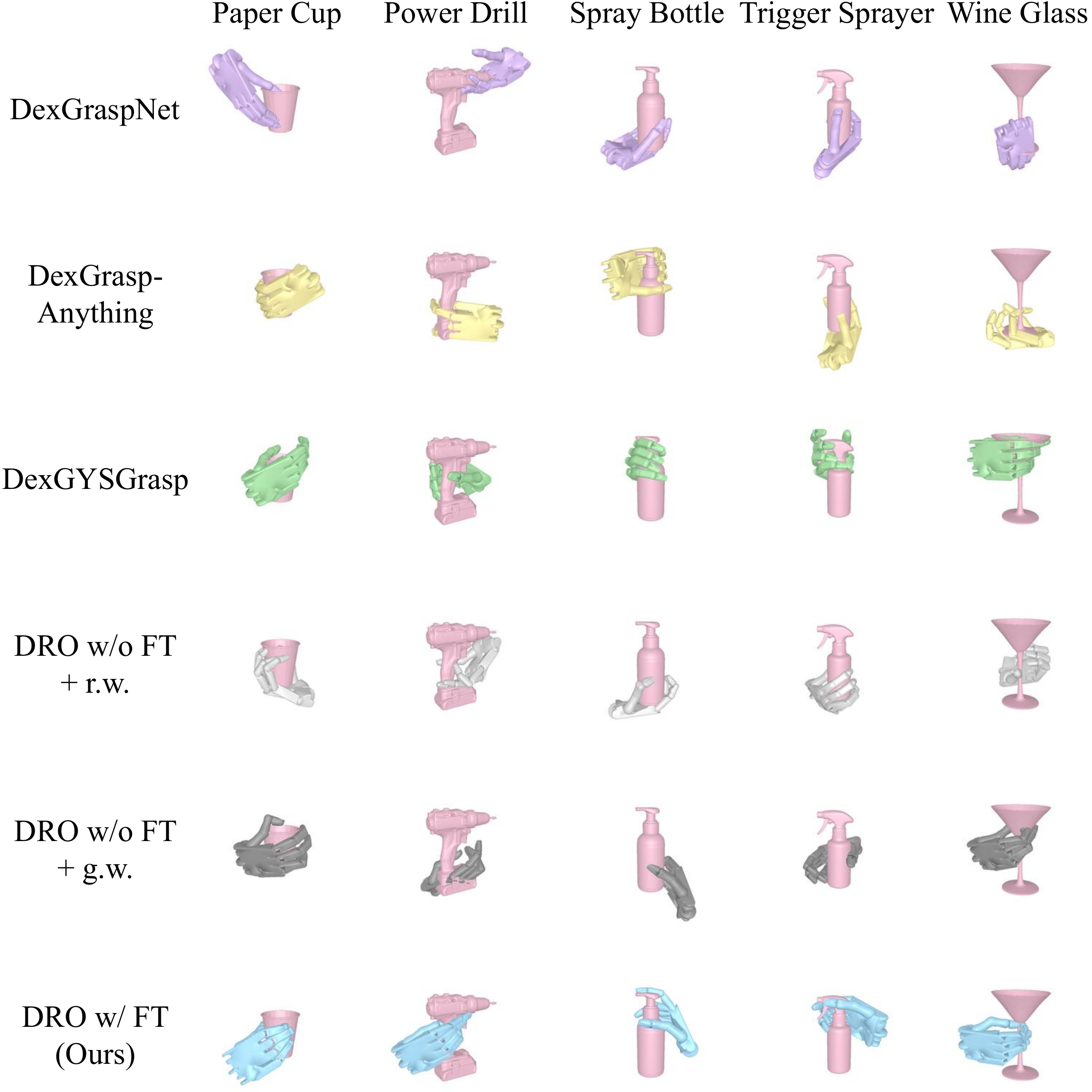}
    \caption{Qualitative comparison of generated grasps across the 5 object categories.  Compared with baselines that often produce stable but semantically generic contacts, DRO fine-tuned with IM-ENGINE data produces grasps that better align with the intended functional regions, such as drill handles, spray triggers, and glass stems.}
    \label{fig:grasp_vis_comparison}
\end{figure}

\subsection{Functionality Criteria}
\label{app:appendix_func_criteria}
In~\Cref{tab:func_grasp_exp2}, we report both the lifting success rate and the functional success rate. We detail the functionality criteria here.
Concretely, a grasp is considered functional only if it succeeds in the simulated lifting test and satisfies the category-specific criteria listed below:

\begin{itemize}[leftmargin=*]
    \item \textbf{Paper cup} The hand must either wrap around the cup body with the thumb opposing the fingers, or approach from above and grip the rim/upper body in a top-down configuration. Both hold styles allow stable transfer without crushing the cup. Grasps that contact only the base, pinch too narrow a region of the body without sufficient coverage, or make only superficial contact that would not support the cup's weight are non-functional.

    \item \textbf{Power drill} The index finger must be placed on or immediately adjacent to the trigger, with the remaining fingers wrapping around the handle grip. Grasps that contact the drill body, chuck, or battery pack without engaging the handle, or that position the index finger away from the trigger, are non-functional.

    \item \textbf{Spray bottle} The index finger must be placed on or immediately adjacent to the pump lever while the remaining fingers stabilize the bottle body. Grasps where the index finger rests on the body rather than the pump region are non-functional.

    \item \textbf{Trigger sprayer} The index finger must be placed on the trigger lever, with the remaining fingers wrapped around the handle. The trigger region is more geometrically distinct than the spray bottle pump, so the criterion is applied strictly: any grasp where the index finger does not contact the trigger is non-functional.

    \item \textbf{Wine glass} The hand must contact the stem or lower bowl in a stable configuration that allows controlled holding. Grasps that contact only the base or that risk tipping the glass are non-functional.
\end{itemize}

\subsection{Visual Results}
\label{app:grasp_visual_results}

\begin{figure}[t]
    \centering
    \includegraphics[width=1.0\textwidth]{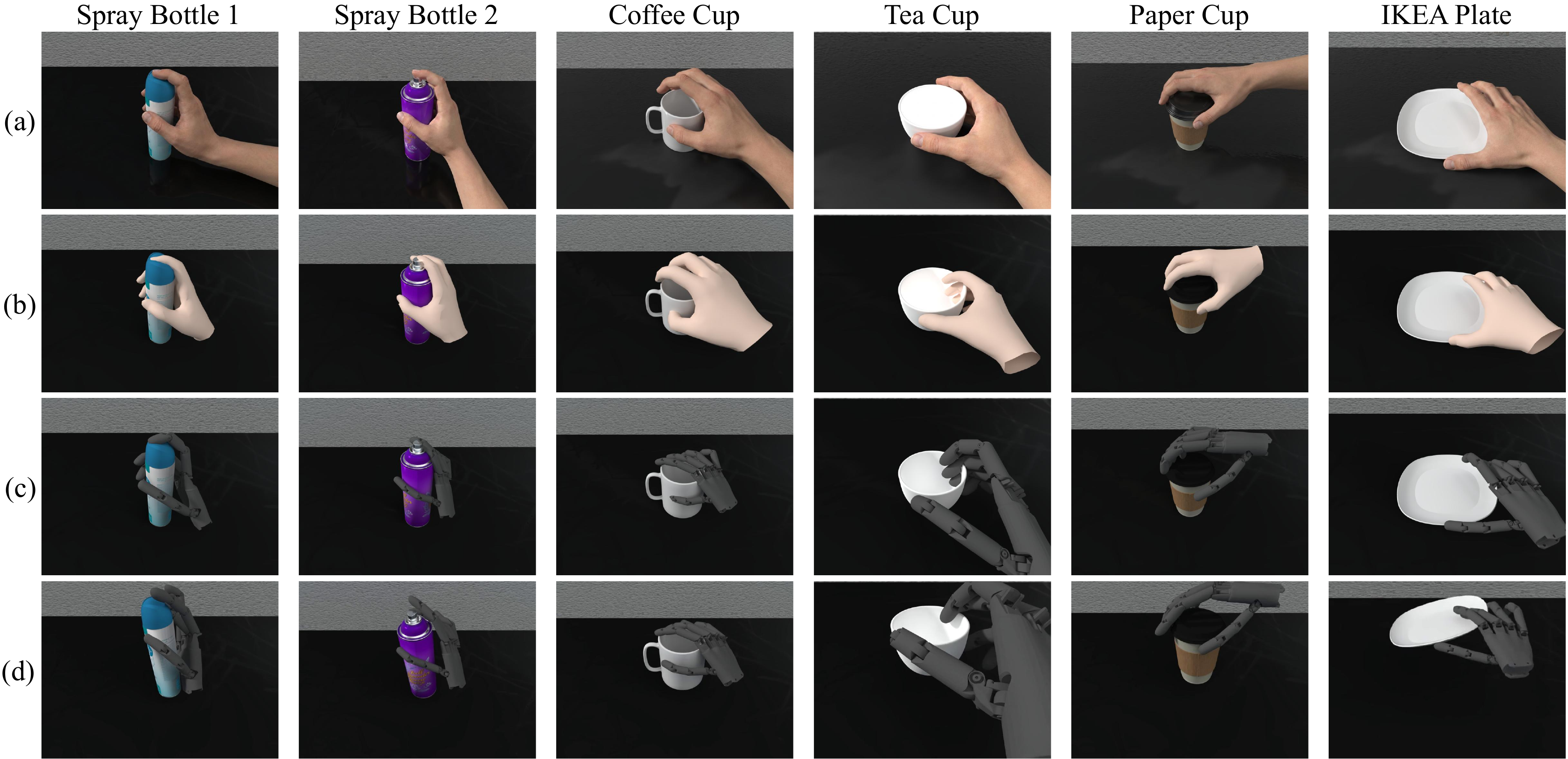}
    \caption{Qualitative results of our grasp-generation pipeline on diverse objects. For each example, we show (a) the edited image, (b) the estimated MANO pose, (c) the retargeted dexterous hand pose, and (d) the physics-validated lifting result.}
    \label{fig:functional_grasping_demo}
\end{figure}

In~\Cref{tab:func_grasp_exp2}, we report the quantitative results on functional grasping. Fig.~\ref{fig:grasp_vis_comparison} provides qualitative examples across the 5 evaluated categories: paper cup, power drill, spray bottle, trigger sprayer, and wine glass. The visual comparison supports the quantitative trend: methods optimized primarily for lifting, such as DexGraspNet~\cite{wang2023dexgraspnet} and DexGrasp-Anything~\cite{zhong2025dexgraspanything}, often find contacts that stabilize the object but miss task-relevant regions. For example, they may grasp the body of a drill or spray bottle without placing the index finger near the trigger, or hold a wine glass in a way that is stable but not functionally appropriate.

Language-conditioned DexGYSGrasp~\cite{wei2024graspasyousay} produces more semantically plausible contacts in some cases, especially when the desired grasp region is visually salient, but it still struggles with precise fingertip placement. The DRO~\cite{wei2025mathcal} ablations show that initialization matters: guided wrist initialization yields grasps that are closer to functional regions than random initialization, but the contacts remain inconsistent without fine-tuning. In contrast, DRO w/ FT (Ours) more consistently places the hand on category-specific functional regions, such as wrapping the paper cup body, grasping the drill handle, contacting the spray and trigger mechanisms, and holding the wine glass near the stem or lower bowl. These qualitative results illustrate why our method improves functional success more than lifting success alone.

Fig.~\ref{fig:functional_grasping_demo} further visualizes intermediate outputs of our pipeline across diverse object categories, including the edited image, estimated MANO hand pose, retargeted robot-hand pose, and final physics-refined grasp. These examples show that our pipeline generalizes across varied object geometries rather than relying on category-specific shape assumptions.

\subsection{Failure Cases}
\label{app:grasp_failure_cases}

\begin{figure}[htp]
    \centering
    \includegraphics[width=\textwidth]{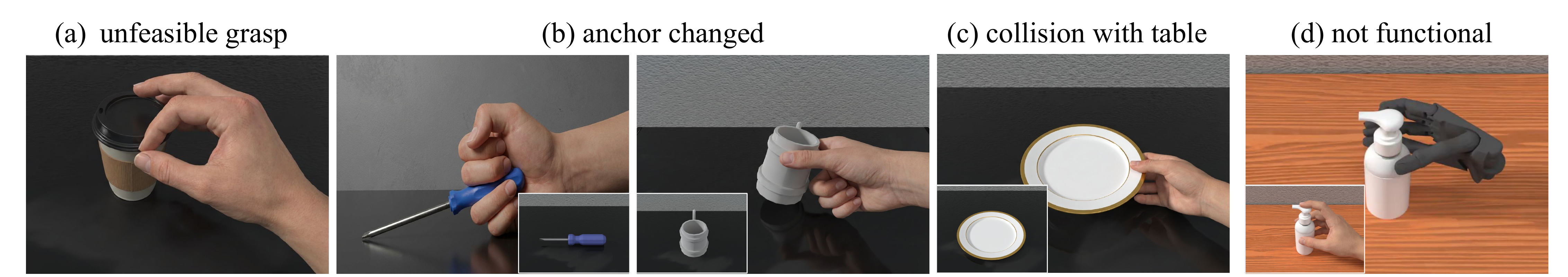}
    \caption{Typical failure modes in functional grasp generation: (a) infeasible grasps, where the edited image does not depict a physically feasible hand-object interaction; (b) anchor changes, where the edited image moves the anchor object despite prompting, breaking depth alignment; (c) table collisions, where the generated hand penetrates the table; and (d) non-functional grasps, where loose contact constraints during search and validation yield stable but functionally incorrect grasps.}
    \label{fig:grasp_failure}
\end{figure}

Fig.~\ref{fig:grasp_failure} summarizes the main failure modes in functional grasp generation. Infeasible edited grasps, anchor-object changes, and table penetrations occur only rarely, but they can still disrupt reconstruction or validation when they appear. Table-penetration failures are partly caused by the pure black table used in scene setup, which can make the table boundary ambiguous for image editing; using a textured table would reduce these cases substantially. The most common failure mode is non-functional grasping: because our search-and-validation stage uses relatively loose contact constraints, it can occasionally find grasps that are physically stable but do not preserve the intended functional contact pattern. The functional proportions are 95\% for paper cup, 77\% for power drill, 71\% for spray bottle, 90\% for trigger sprayer, and 98\% for wine glass. 

\section{Goal Generation}
\label{app:goal_generation}

\begin{figure}[t]
    \centering
    \includegraphics[width=1.0\textwidth]{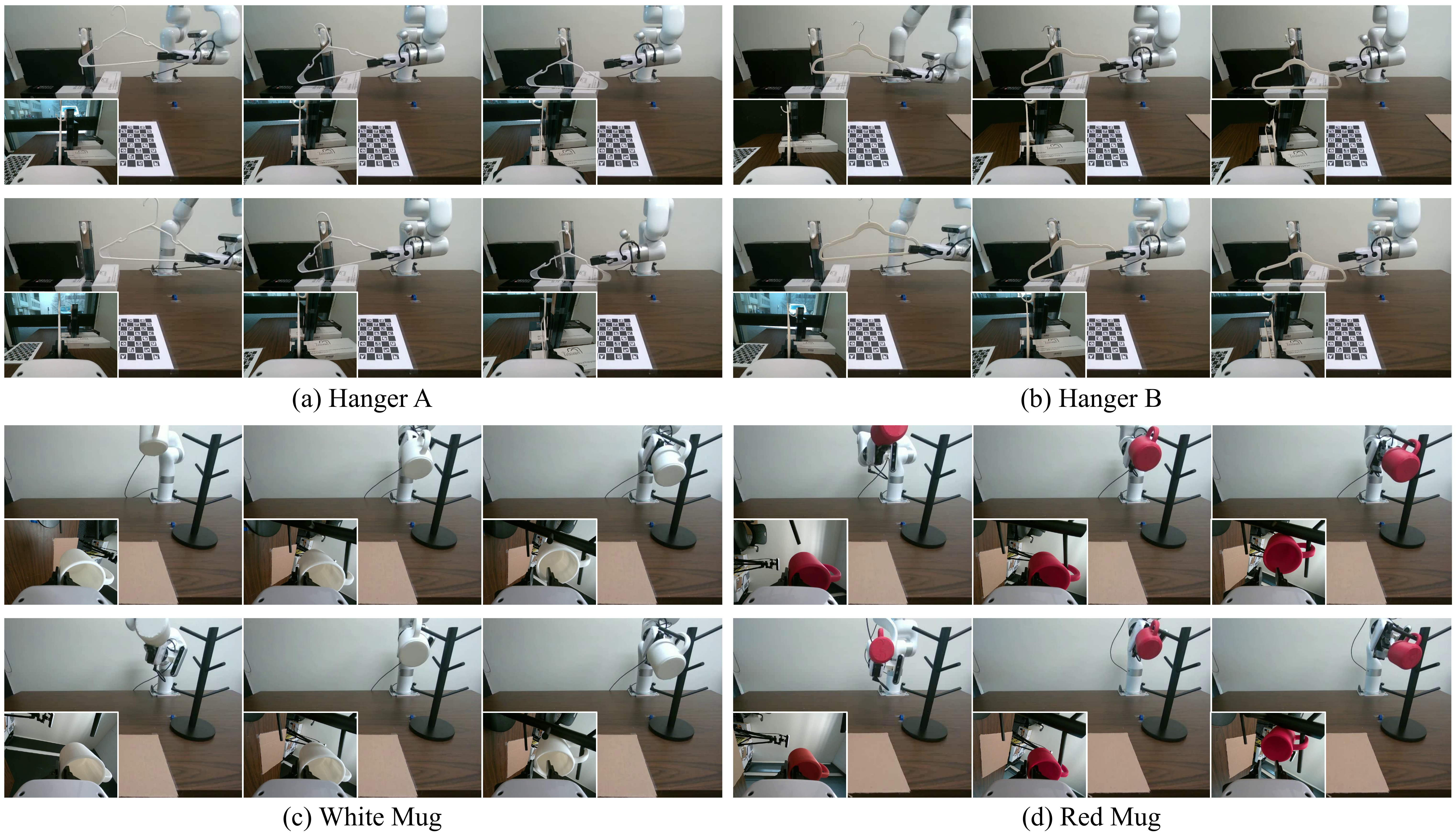}
    \caption{Additional real-world goal-generation results. We supplement more results obtained by the fine-tuned SmolVLA policies described in~\Cref{sec:real_world_goal_gen}.}
    \label{fig:real-world_goal-gen_additional}
\end{figure}

\subsection{Method Details}
\label{app:goal_physics_validation}

\paragraph{Goal-generation image-editing prompts.}
For goal-state generation, we use the following prompt template:
\begin{lstlisting}[basicstyle=\ttfamily\small,breaklines=true]
Please generate an image to {task_description}. Please keep the same camera viewpoint and lighting. Make sure to move/reposition the target objects as described instead of creating new ones. The object should be removed from its original place. Maintain realistic physics and object proportions.
\end{lstlisting}
The task descriptions are case-specific: \textbf{Mug}: hang the mug on the mug tree; \textbf{Dish 1}: place the plate on the rack, with the rim wedged between the two rails; \textbf{Dish 2}: place the plate on the rack, with the rim wedged between the two rails; \textbf{Hanger}: hang the hanger on the hooks on the wall; \textbf{Tube}: insert the tube to the correct position on the rack; \textbf{Scissor}: hang the scissors on the holder bar; \textbf{Wine 1}: place the wine bottle on the wine rack; and \textbf{Wine 2}: place the wine bottle in the wine rack.

\paragraph{Goal-pose estimation.}
\label{app:goal_pose_estimation}
Given the edited goal image, target-object mask, aligned depth, camera intrinsics, camera pose, and target mesh, we estimate a world-frame 6-DoF pose for the target object. We first sample 48 object rotations on $SO(3)$ with a stratified Hopf-fibration grid and render the target mesh under each candidate rotation using the goal-image intrinsics, producing template images and template depth point clouds with known template object-to-camera transforms $T_{co}^{(m)}$. DINOv3 features provide dense correspondences between each rendered template and the edited target object. For each accepted match, the template pixel is back-projected through the rendered template depth to a template-camera point $a_{c,i}$ and transformed to an object-frame point $A_i=(T_{co}^{(m)})^{-1}a_{c,i}$; the corresponding goal-image pixel is back-projected through the aligned depth map to a goal-camera point $b_{c,i}$ and transformed to a world-frame point $B_i=T_{wc}b_{c,i}$. We keep only depth-consistent correspondences and solve the rigid 3D registration
\begin{equation}
(R^{\star}, t^{\star}) = \arg\min_{R\in SO(3),\,t\in\mathbf{R}^3} \sum_i \left\|B_i - (R A_i + t)\right\|^2
\end{equation}
with the closed-form Kabsch/Procrustes solution. With this convention, the recovered transform maps target-object coordinates to the simulator world frame, i.e., $T_{wo}^{\star}=[R^{\star}\mid t^{\star}]$. Let $\bar{A}=N^{-1}\sum_i A_i$ and $\bar{B}=N^{-1}\sum_i B_i$ denote the centroids, and let $\tilde{A}_i=A_i-\bar{A}$ and $\tilde{B}_i=B_i-\bar{B}$. For any fixed rotation $R$, the optimal translation is
\begin{equation}
t(R)=\bar{B}-R\bar{A},
\end{equation}
so the remaining problem is to align the centered point sets. We compute the cross-covariance $H=\sum_i \tilde{A}_i\tilde{B}_i^{\top}$ and its SVD $H=U\Sigma V^{\top}$. The rotation is projected onto $SO(3)$ with the reflection guard
\begin{equation}
R^{\star}=V\,\mathrm{diag}\left(1,1,\det(VU^{\top})\right)U^{\top},
\end{equation}
This prevents the raw SVD solution from returning an improper rotation. We then set $t^{\star}=t(R^{\star})$ and score each template by correspondence recall and registration residual,
\begin{equation}
e=\sqrt{\frac{1}{N}\sum_i \left\|B_i-(R^{\star}A_i+t^{\star})\right\|^2},
\end{equation}
and use the best Procrustes pose as the initial transform for ICP.

The ICP refinement solves the same rigid alignment objective but replaces fixed DINOv3 correspondences with nearest-neighbor correspondences to a denser goal point cloud. We construct this cloud by back-projecting all pixels in the edited target-object mask through the aligned depth map, transform the points into the world frame, and remove isolated outliers with DBSCAN. Starting from the Procrustes transform $T_{\mathrm{init}}$, point-to-point ICP alternates between nearest-neighbor assignment
\begin{equation}
c(i)=\arg\min_j \left\|Q_j-(R P_i+t)\right\|^2
\end{equation}
and rigid alignment
\begin{equation}
(R,t)=\arg\min_{R\in SO(3),\,t\in\mathbf{R}^3}\sum_i\left\|Q_{c(i)}-(R P_i+t)\right\|^2,
\end{equation}
where $P_i$ are source mesh points in the target-object frame, $Q_j$ are depth-derived goal points in the world frame, and $(R,t)$ is always an object-to-world transform. We run ICP in two stages: first on the sparse matched template points to correct small orientation errors, then on 5000 points uniformly sampled from the full target mesh to let the complete object surface drive the final alignment. The final object-to-world ICP pose is passed to physics validation when available.

\paragraph{Physics validation.}
For goal-state generation, the recovered object pose is refined by sampling local perturbations around $(R,t)$ and simulating them with the fixed anchor, table, and optional wall. We reject candidates that begin in collision with the anchor or environment. The remaining candidates are stepped forward under gravity and contact; a pose is accepted only if its translation and rotation drift stay below thresholds after simulation. This catches visually plausible but physically invalid edits, including floating placements, penetrations, placements that exploit inaccurate collision geometry around rack gaps, and unstable hanging relations.

\paragraph{Trajectory generation.}
The refined target pose is a constrained terminal relation rather than a simple placement. A standard pick-and-place routine often fails for our selected tasks because the object must enter narrow free space near the anchor, such as threading a mug onto a mug-tree branch or inserting a plate into a rack slot. We therefore first compute a kinematic escaping path from the validated goal pose with a model predictive control (MPC) procedure in object-pose space. At each MPC step, a predictive-sampling kernel proposes a batch of short-horizon action sequences, where each action directly applies a small translation and rotation to the target object. We roll out these candidates with a kinematic geometry model, reject sequences that collide with the anchor, table, or wall, execute the first action of the best collision-free sequence, and then replan from the updated pose. Because this stage only applies pose updates and collision checks, rather than simulating contact dynamics, it quickly finds a collision-free path that moves the object out of the constraint. Reversing the escape path yields a kinematic insertion path from a distant free-space pose into the target relation. This reversed path serves as the warm start for a forward MPC search in simulation; without this initialization, the forward search rarely discovers the narrow constrained insertion trajectory.

\subsection{Additional Baseline Analysis}
\label{app:goal_baseline_analysis}

\begin{figure}[htp]
    \centering
    \includegraphics[width=\textwidth]{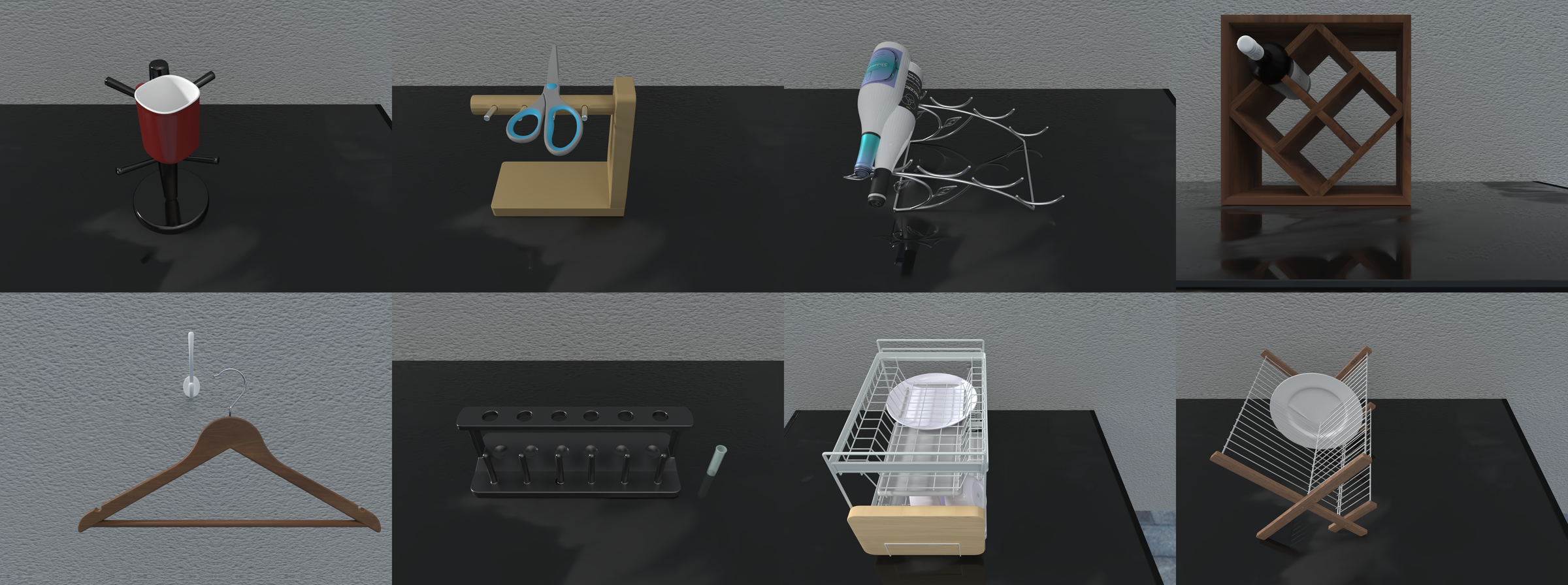}
    \caption{Qualitative results on goal-state generation for Molmo-2D Baseline.}
    \label{fig:molmo_qual}
\end{figure}

\begin{figure}[htp]
    \centering
    \includegraphics[width=\textwidth]{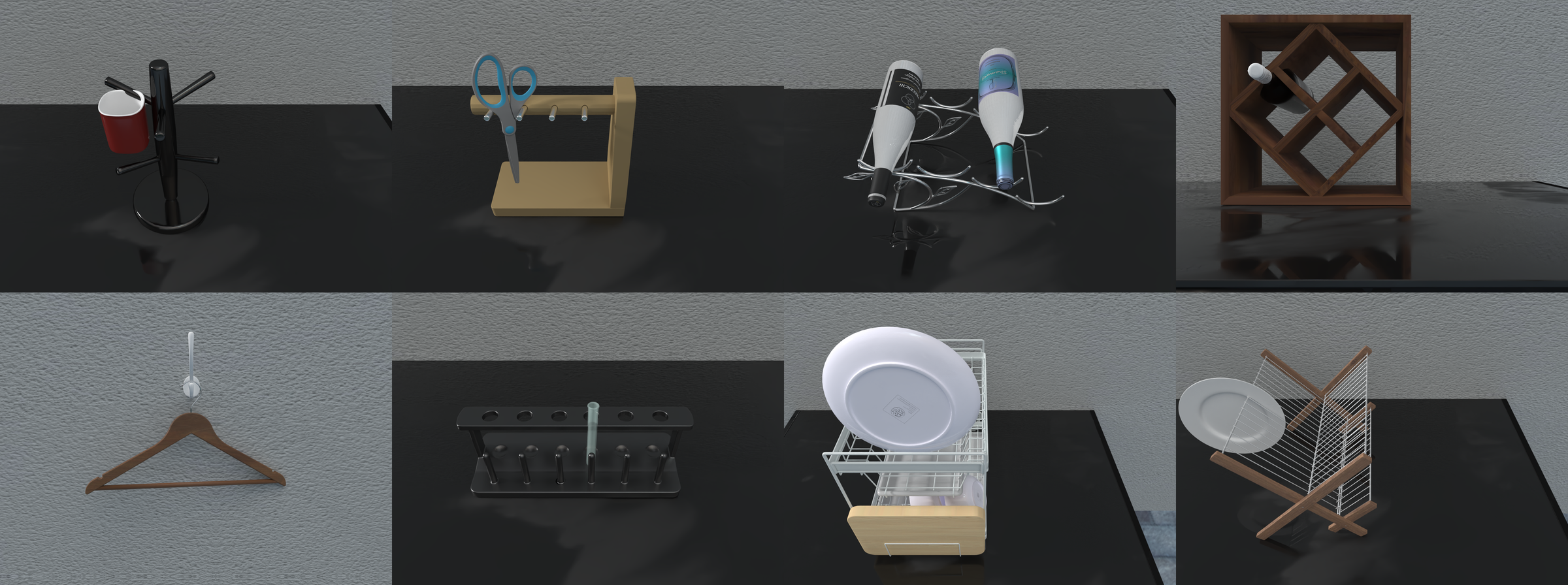}
    \caption{Qualitative results on goal-state generation for VLM-6DoF (5 iterations) Baseline.}
    \label{fig:vlm5iter_qual}
\end{figure}

\begin{figure}[htp]
    \centering
    \includegraphics[width=\textwidth]{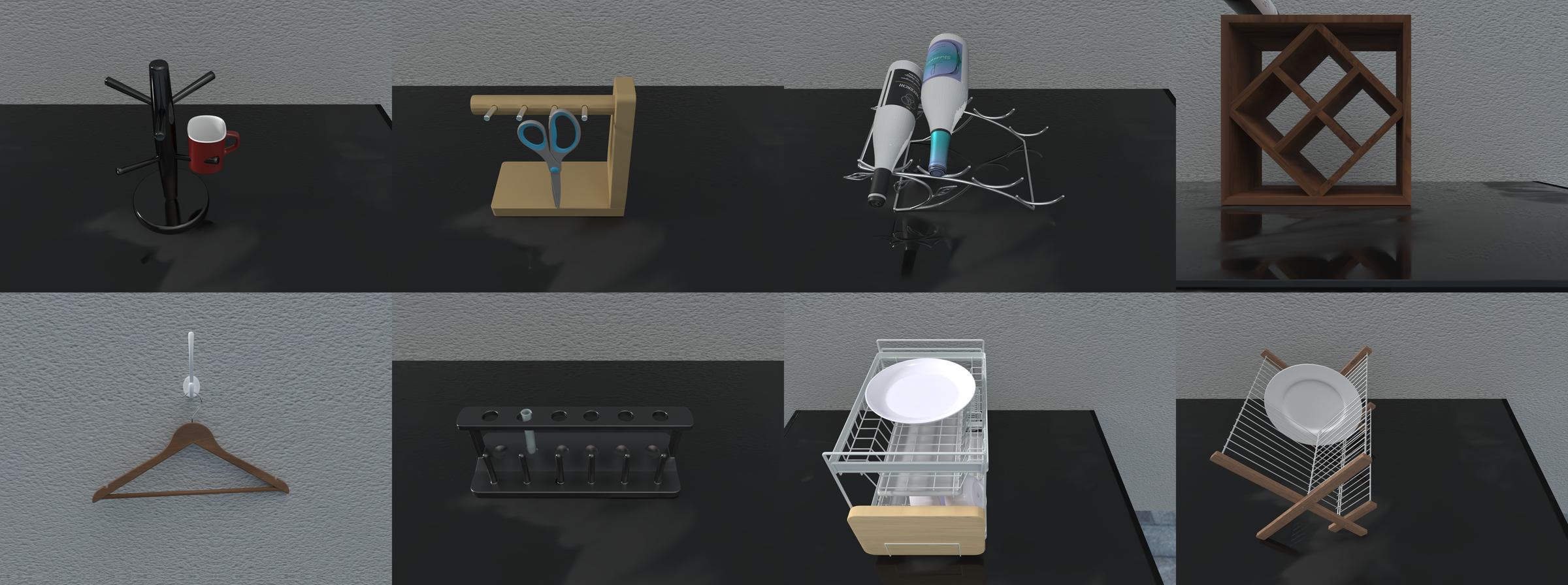}
    \caption{Qualitative results on goal-state generation for VLM-6DoF (10 iterations) Baseline.}
    \label{fig:vlm10iter_qual}
\end{figure}

\begin{figure}[htp]
    \centering
    \includegraphics[width=\textwidth]{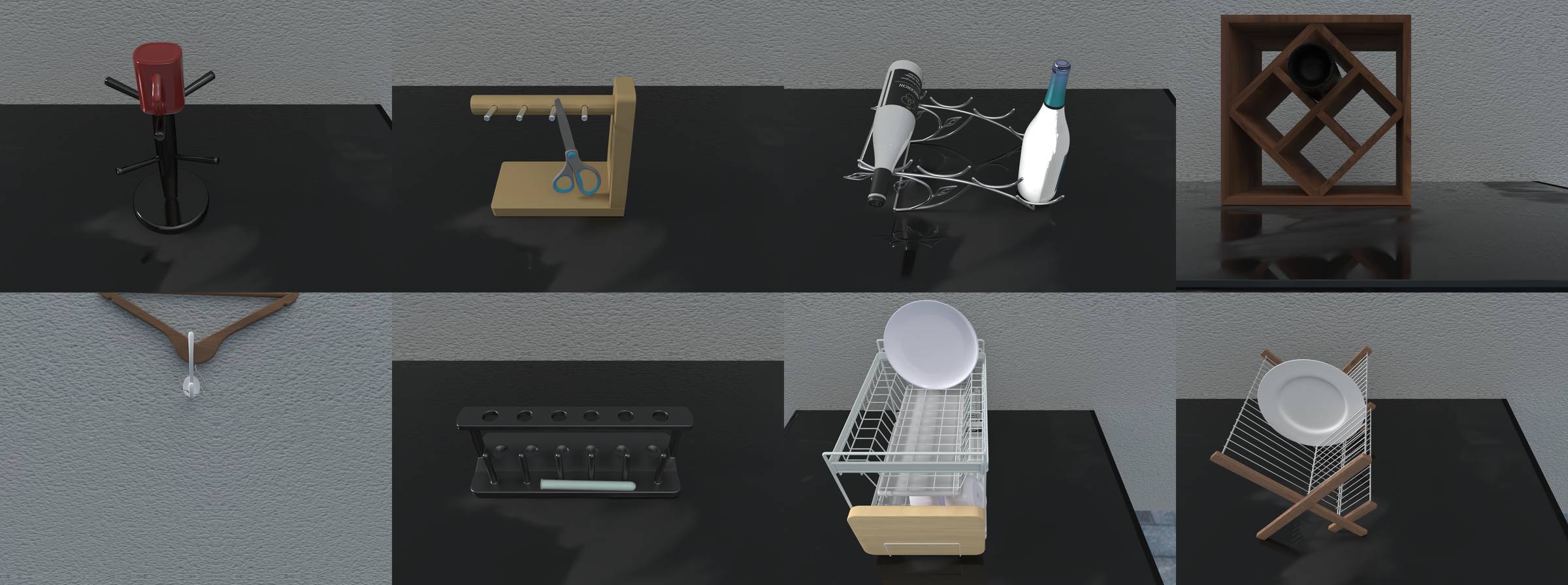}
    \caption{Qualitative results on goal-state generation for Physcensis Baseline.}
    \label{fig:physcensis_qual}
\end{figure}

In~\Cref{sec:exp_goal_gen}, we briefly introduced three baselines we considered in our goal-state generation experiment. Here, we supplement the implementation details of these baselines.

\begin{itemize}[leftmargin=*]
    \item \textbf{Molmo-2D} This baseline utilizes Molmo~\cite{deitke2025molmo}, an open-source VLM with 2D pointing specialty, to identify a feasible pixel based on a task description and a rendered image of the initial state. It subsequently lifts the 2D point into 3D world coordinates using the rendered depth and calibrated camera parameters. The object center is given by the lifted 3D coordinate while the rotation is fixed. The object pose is then validated through the physics refinement and used as the goal pose. 
    \item \textbf{VLM-6DoF} This baseline casts goal generation as iterative VLM prompting. In each step, the current scene is rendered from a fixed viewpoint and sent to the VLM along with the task description and 6-DoF poses for both the target and the container, as well as their sizes. The VLM then provides a new target pose, which is used in the next iteration. After several iterations (we tested both 5 and 10), the final accepted pose is validated through the physics refinement, and this pose is finally used as the goal. We used \texttt{gemini-3.1-pro-preview} in our experiment.
    \item \textbf{PhyScensis} This baseline~\cite{wang2026physcensis} uses an occupancy-grid-based heuristic search pipeline to find 3D object poses for arrangement-style placement tasks, followed by physics validation. Given a fixed rotation, it generates multiple feasible placements; therefore, we sample five feasible cases for each of several candidate rotations and use a VLM agent to select the most task-consistent one.
\end{itemize}

Figs.~\ref{fig:molmo_qual}, \ref{fig:vlm5iter_qual}, \ref{fig:vlm10iter_qual}, and~\ref{fig:physcensis_qual} provide qualitative comparisons for the goal-generation baselines. Molmo-2D often identifies a plausible image-space location, but lifting a single 2D point provides only a coarse translation and does not infer the orientation needed for constrained relations. As a result, the target object is frequently placed near the anchor but fails to realize the intended relation, such as hanging, insertion, or fitting.

The VLM-6DoF baseline improves when allowed more iterations, as shown by the difference between Figs.~\ref{fig:vlm5iter_qual} and~\ref{fig:vlm10iter_qual}. Additional rounds can correct some gross placement errors and move the object closer to the desired region. However, the generated poses remain noisy in metric 3D space, and small translation or rotation errors are enough to break contact-rich tasks with narrow clearances. This explains why the 10-iteration variant improves over the 5-iteration variant but still remains far below our method quantitatively.

PhyScensis can generate physically plausible placements for simpler support-style tasks, but it relies on geometric search and post-hoc selection rather than a direct semantic proposal. As a result, it can miss constrained object-object relations that require a specific orientation or contact configuration, such as insertion, hanging, or fitting. 

In contrast, our image-editing stage proposes a semantically meaningful final configuration before metric pose recovery. This makes the downstream search more targeted: pose estimation and physics validation only need to refine a visually specified relation rather than discover it from scratch. The comparison highlights that physical feasibility alone is insufficient for these tasks; successful goal generation also requires preserving the intended semantic relation between the target and anchor objects.

\subsection{Failure Cases}
\label{app:goal_failure_cases}

\begin{figure}[htp]
    \centering
    \includegraphics[width=\textwidth]{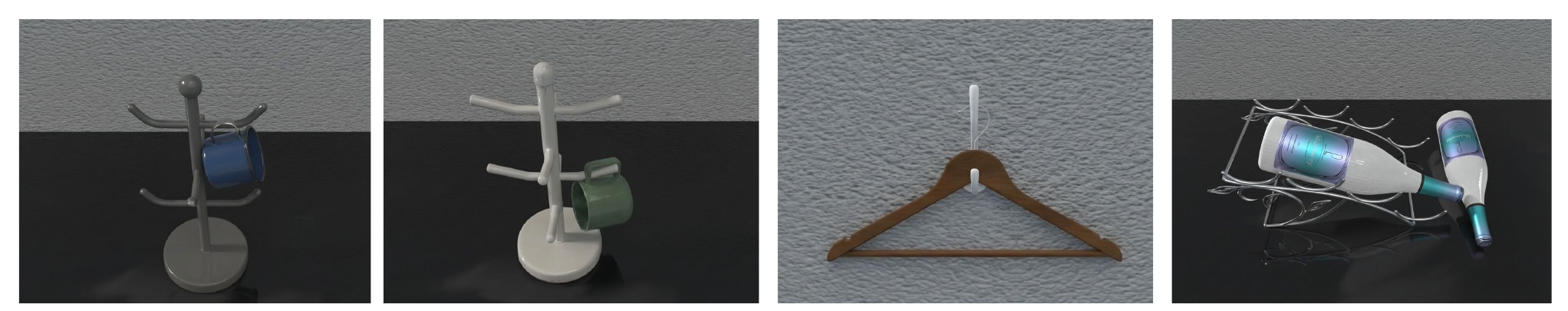}
    \caption{Failure cases in goal-state generation. Most failures come from image editing, where the edited image can produce infeasible goals or alter the anchor object.}
    \label{fig:goal_failure}
\end{figure}

Fig.~\ref{fig:goal_failure} shows representative failure cases of our goal-generation pipeline. Most failures arise from the image-editing stage: the edited image can sometimes specify an infeasible goal state or change the anchor object, making it difficult to recover a physically valid pose that still satisfies the intended semantic relation.

Another limitation is pose estimation. When the target object has approximate symmetries, recovering the exact rotation can be ambiguous. For example, in Fig.~\ref{fig:demo_grid_appendix}, the tube case is estimated and validated upside down. We still count this as a success because the resulting goal is largely correct, physically feasible, and semantically reasonable.

\begin{figure}[t]
    \centering
    \includegraphics[width=\textwidth]{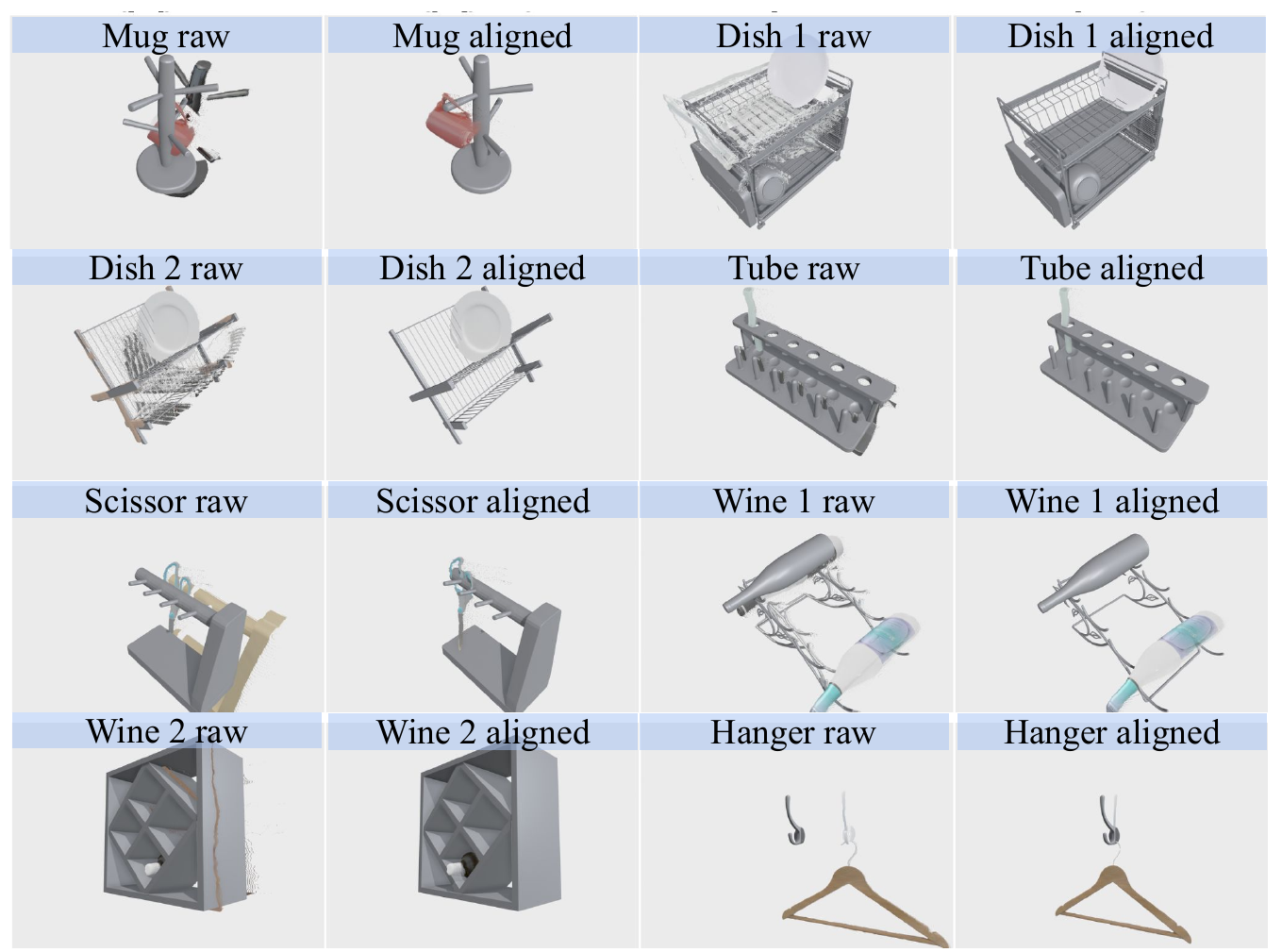}
    \caption{Additional qualitative results on anchor-based point cloud alignment.}
    \label{fig:pcd_alignment_grid_appendix}
\end{figure}

\subsection{Additional Goal-State Generation Visualizations}
\label{app:goal_visualizations}

\begin{figure}[t]
    \centering
    \includegraphics[width=.85\textwidth]{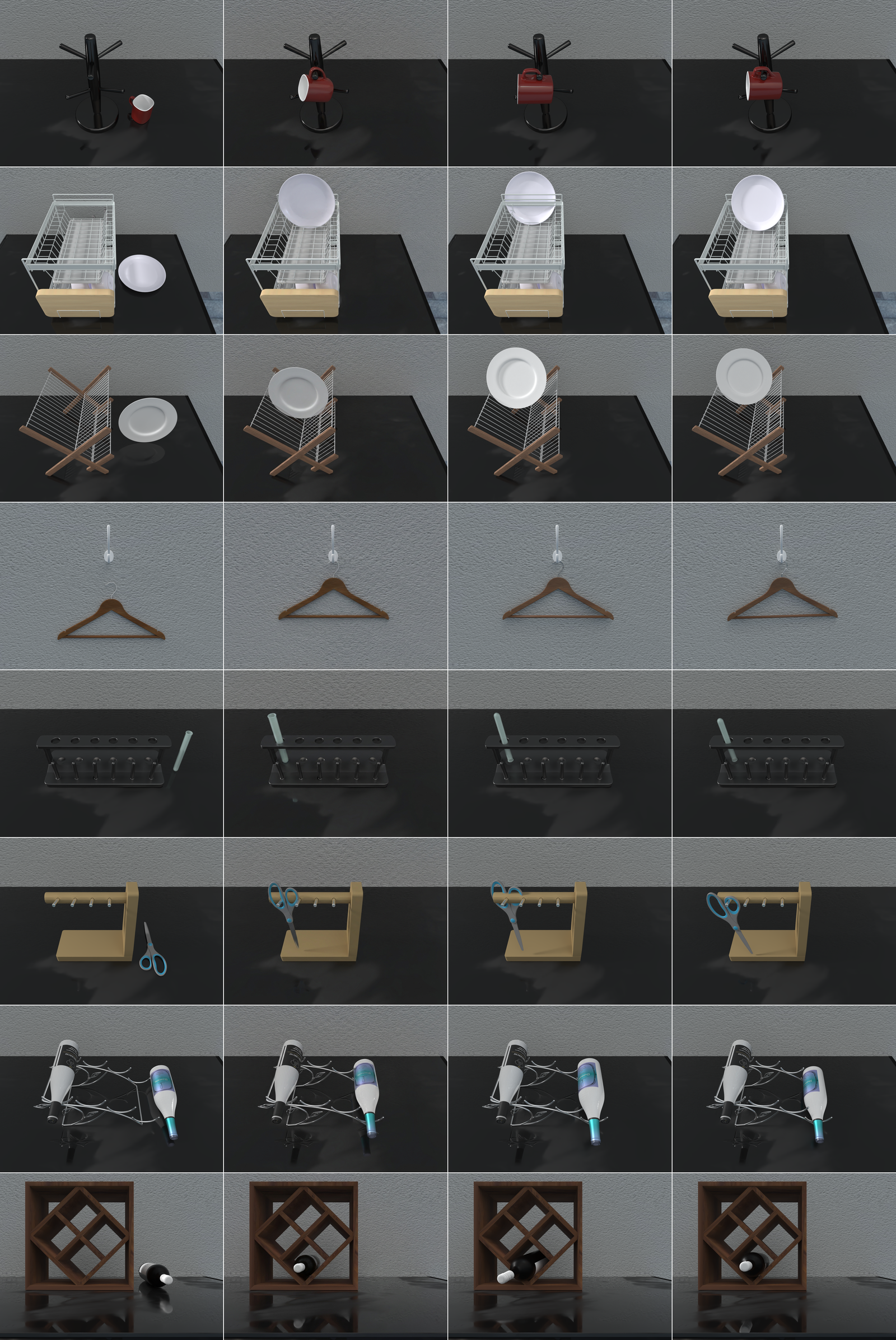}
    \caption{Additional qualitative results for goal-state generation. For each example, the first column shows the original image, the second column shows the edited image, the third column shows the estimated pose, and the final column shows the physics-validated goal pose.}
    \label{fig:demo_grid_appendix}
\end{figure}

Fig.~\ref{fig:demo_grid_appendix} shows additional qualitative examples of our goal-state generation pipeline. The image-editing stage largely preserves the anchor object unchanged while generating reasonable and semantically correct target goals. Although the edited result is not always perfectly accurate, it provides a sufficiently informative visual goal for downstream pose recovery. The initially estimated pose may also be imperfect, but by applying small pose jitters followed by physics validation, we can recover a precise and physically plausible goal pose.

Fig.~\ref{fig:pcd_alignment_grid_appendix} visualizes the effect of our depth-alignment procedure. Off-the-shelf depth estimation does not directly provide sufficiently accurate absolute depth for these contact-rich tasks, but after our alignment step, the resulting point cloud is accurate enough for downstream pose estimation and validation. We visualize this by placing the target object's point cloud in the world frame together with the anchor object mesh at its world-frame pose. In the mug and hanger cases, the raw depth projection is clearly displaced far from the correct location, while our alignment corrects the projection and produces an accurate enough point cloud for subsequent processing.

\end{document}